\documentclass[11pt,a4paper]{article}
\usepackage[hyperref]{acl2021}
\usepackage{times}
\usepackage[utf8]{inputenc}
\usepackage{latexsym}
\usepackage{booktabs}
\usepackage{multicol}
\usepackage{graphicx}
\usepackage{amsmath}
\usepackage{amsfonts}
\usepackage{inconsolata}
\usepackage{framed}
\usepackage{linguex}
\usepackage{xcolor}
\usepackage{pgf}
\usepackage{collcell}

\usepackage{microtype}
\def\aclpaperid{2939} %  Enter the acl Paper ID here

\title{Causal Analysis of Syntactic Agreement Mechanisms\\in Neural Language Models}

\aclfinalcopy

\author{
Matthew Finlayson\Thanks{ Equal contribution.} \\ Harvard University \\ Cambridge, MA \\ \texttt{mattbnfin@gmail.com}
\And Aaron Mueller\footnotemark[1] \\ Johns Hopkins University \\ Baltimore, MD \\ \texttt{amueller@jhu.edu}
\And Sebastian Gehrmann \\ Google Research \\ New York, NY \\ \texttt{gehrmann@google.com} 
\AND Stuart Shieber \\ Harvard University \\ Cambridge, MA \\ \texttt{shieber@seas.harvard.edu}
\And Tal Linzen\Thanks{ Work done while visiting Google Research.} \\ New York University \\ New York, NY \\ \texttt{linzen@nyu.edu}
\And Yonatan Belinkov\Thanks{ Supported by the Viterbi Fellowship in the Center for Computer Engineering at the Technion.} \\ Technion -- IIT \\ Haifa, Israel \\ \texttt{belinkov@technion.ac.il}
}

\date{}

\begin{document}
\maketitle
\begin{abstract}

Targeted syntactic evaluations have demonstrated the ability of language models to perform subject-verb agreement given difficult contexts. 
To elucidate the mechanisms by which the models accomplish this behavior, this study applies causal mediation analysis to pre-trained neural language models.
We investigate the magnitude of models' preferences for grammatical inflections, as well as whether neurons process subject-verb agreement similarly across sentences with different syntactic structures.
We uncover similarities and differences across architectures and model sizes---notably, that larger models do not necessarily learn stronger preferences.
We also observe two distinct mechanisms for producing subject-verb agreement depending on the syntactic structure of the input sentence.
Finally, we find that language models rely on similar sets of neurons when given sentences with similar syntactic structure.

\end{abstract}

\setlength{\Exlabelwidth}{0.25em}
\setlength{\SubExleftmargin}{1.35em}

\section{Introduction}
Targeted syntactic evaluations have shown that neural language models (LMs) are able to predict the correct token from a set of grammatically minimally different continuations with high accuracy, even in difficult contexts \citep{dupouxlinzen16,gulordavacolorless18}, for constructions such as subject-verb agreement \citep{martylinzen19}, filler-gap dependencies \citep{wilcox18gap}, and reflexive anaphora \citep{marvinlinzen18}.

As an illustration of the targeted syntactic evaluation paradigm, consider the following example, which demonstrates subject-verb agreement across an agreement attractor. Here, a model using a linear analysis (i.e., inflecting based on the most recent noun) would choose the ungrammatical inflection, while a model using a hierarchical analysis would choose the grammatical inflection:

\ex.The key to the cabinets \underline{is}/*\underline{are} next to the coins.

While we have a reasonable understanding of the generally correct \emph{behavior} of LMs in such contexts, the mechanisms that underlie models' sensitivity to syntactic agreement are still not well understood. Recent work has performed causal analyses of syntactic agreement units in LSTM \citep{lstm}-based LMs \citep{lakretz2019emergence, lu2020influence} or causal analyses of LSTM hidden representations' impact on syntactic agreement \citep{giulianelli2018under}, but the agreement mechanisms of Transformer-based LMs have not been as extensively investigated. Transformer-based LMs' syntactic generalization abilities are superior to those of LSTMs \citep{hu2020systematic}, which makes Transformer-based models enticing candidates for further analysis.

We apply the behavioral-structural method of causal mediation analysis~\citep{pearl2001direct} to investigate syntactic agreement in Transformers, following the approach used by \citet{vig2020causal} for interpreting gender bias in pre-trained English LMs. This method allows us to implicate specific model components in the observed behavior of a model. If we view a neural LM as a causal graph proceeding from inputs to outputs, we can view each model component (e.g., a neuron) as a mediator. We measure the contribution of a mediator to the observed output behavior by performing controlled \emph{interventions} on input sentences and observing how they change the probabilities of continuation pairs. We focus primarily on GPT-2 \citep{gpt2}, although we also analyze TransformerXL~\citep{transformerxl} and XLNet~\citep{xlnet}.

We find that both GPT-2 and Transformer-XL use two distinct mechanisms to accomplish subject-verb agreement, one of which is active only when the subject and verb are adjacent. Conversely, \mbox{XLNet} uses one unified mechanism across syntactic structures. Even though larger models assign a higher probability to the correct inflection more often, this does not necessarily translate to a larger margin between the probability of the correct and incorrect options.
Additionally, in larger models, agreement mechanisms are similar to those in smaller models, but are more distributed across layers. Finally, we find that the most important neurons for agreement are shared across different structures to various extents, and that the degree of neuron overlap matches well with human intuitions of syntactic similarity between structures.

\section{Related Work}
\subsection{Targeted Syntactic Evaluation}
Many recent studies have treated neural LMs and contextualized word prediction models---primarily LSTM LMs~\citep{sundermeyer2012lstm}, GPT-2~\citep{gpt2}, and BERT~\citep{bert}---as psycholinguistic subjects to be studied behaviorally~\citep{dupouxlinzen16,gulordavacolorless18,bertgoldberg19}.
Some have studied whether models prefer grammatical completions in subject-verb agreement contexts~\citep{marvinlinzen18,martylinzen19,bertgoldberg19,mueller2020cross, LAKRETZ2021104699, futrell2019neural}, as well as in filler-gap dependencies~\citep{wilcox18gap,wilcox2019structural}. These are based on the approach of \citet{dupouxlinzen16}, where a model's ability to syntactically generalize is measured by its ability to choose the correct inflection in difficult structural contexts instantiated by tokens that the model has not seen together during training. In other words, this approach tests whether the model assigns the correct inflection a higher probability than an incorrect inflection given the same context.
This approach investigates the output behavior of the model, but does not inform one of how the model does this or which components are responsible for the observed behavior.

\subsection{Probing}
A separate line of analysis work has investigated representations associated with syntactic dependencies by defining a family of functions (probes) that map from model representations to some phenomenon that those representations are expected to encode.
For instance, several studies have mapped LM representations to either independent syntactic dependencies \citep{belinkov:2018:phdthesis,liu-etal-2019-linguistic,tenney2018what} or full dependency parses \citep{hewitt2019structural,chi2020mbert_probe} as a proxy for discovering latent syntactic knowledge within the model.
Most related, \citet{giulianelli2018under} use probes to investigate how LSTMs handle agreement.

Probing is more difficult to interpret than behavioral approaches because the addition of a trained classifier introduces confounds~\citep{hewitt2019selectivity}: most notably, whether the probe maps from model representations to the desired output, or learns the task itself. Probes also only give correlational evidence, rather than causal evidence \citep{belinkov2019analysis}. See \citet{belinkov2021probing} for a review of the shortcomings of probes.

\subsection{Causal Mediation Analysis}
Causal inference methods study the change in a response variable following an intervention; for example, how do health outcomes change after a patient stops consuming nicotine products? Causal mediation analysis \citep{robins1992identifiability,pearl2001direct,robins2003causaldag} focuses on the role of a mediator in explaining the effect of a treatment on outcomes. For example, if a patient stops using tobacco, are health outcomes mediated by the initial method of nicotine delivery (e.g., smoking tobacco vs.\ patches vs.\ nicotine gum)? 

This approach lends itself well to interpreting NLP models, as we can view a deep neural network as a graphical model from input to output via mediators, where mediators can be individual components (e.g., neurons). For LMs, the intervention is a change to the input sentence, and the outcome is a function of the probabilities of a set of continuations. 

This approach for interpreting NLP models was introduced by \citet{vig2020causal}, who implicate specific neurons and attention heads in mediating gender bias in various pre-trained LMs.
While one ideally expects equal preferences for male and female completions given gender-ambiguous contexts (for example, given the prompt $u$ ``The nurse said that'', we want $p(\text{she}|u) \approx p(\text{he}|u)$),
this is not the case for subject-verb agreement, where we expect very strong preferences for grammatically correct completions over incorrect completions. 

\section{Experimental Setup}
\subsection{Data}
First, we define prompts $\mathbf{u}$. These prompts are a set of left contexts (beginnings of sentences), generated from a vocabulary and a set of templates developed by \citet{lakretz2019emergence}. We expand the vocabulary with additional tokens, and add relative clause (RC) templates. We opt to synthetically generate prompts rather than sample from a corpus to control for the potential confound of token collocations in the training set.
We use prompts from six syntactic structures; an example of each may be found in Figure~\ref{fig:examples}. For each structure, we randomly sample 300 prompts from all possible noun-verb combinations. Our dataset, code, and random seeds are available on Github.\footnote{\url{https://github.com/mattf1n/lm-intervention}}

\newlength{\vs}
\setlength{\vs}{0.4\baselineskip}
\begin{figure}[t]
        \rule{\columnwidth}{1pt}
        \centering
        \resizebox{0.95\columnwidth}{!}{
        
        \begin{minipage}{\columnwidth}
            \raggedright
            \vspace{0.2cm}
         \textit{Simple Agreement}:\\
         The \textcolor{blue}{athlete} \underline{\textcolor{blue}{confuses}/*\textcolor{red}{confuse}}
         
        \vspace{\vs}\textit{Within Object Relative Clause:}\\
        The \textcolor{red}{friend} (that) the \textcolor{blue}{lawyers} \underline{*\textcolor{red}{likes}/\textcolor{blue}{like}}
        
        \rule{0.9\columnwidth}{0.6pt}
        
        \vspace{\vs}\textit{Across One Distractor}:\\
        The \textcolor{blue}{kids} gently \underline{*\textcolor{red}{admires}/\textcolor{blue}{admire}}
        
        \vspace{\vs}\textit{Across Two Distractors}:\\
        The \textcolor{blue}{father} openly and deliberately \underline{\textcolor{blue}{avoids}/*\textcolor{red}{avoid}}
        
        \rule{0.9\columnwidth}{0.6pt}
         
        \vspace{\vs}\textit{Across Prepositional Phrase}:\\
        The \textcolor{blue}{mother} behind the \textcolor{red}{cars}
        \underline{\textcolor{blue}{approves}/*\textcolor{red}{approve}}

        \vspace{\vs}\textit{Across Object Relative Clause:}\\
        The \textcolor{blue}{farmer} (that) the \textcolor{red}{parents} love \underline{\textcolor{blue}{confuses}/*\textcolor{red}{confuse}}
        \vspace{0.2cm}
        \end{minipage}
    }
        \rule{\columnwidth}{1pt}
    \caption{Syntactic structures used in this study. Ungrammatical forms are marked with asterisks. Target subjects and their agreeing verb inflections are shown in \textcolor{blue}{blue}, while attractors and their agreeing inflections are shown in \textcolor{red}{red}.}
    \label{fig:examples}
\end{figure}

In the `simple agreement' and `within RC' constructions, there is no separation between the target subject and verb. The `across one distractor' and `across two distractors' structures test the effect of placing one or two adverbs between the subject and verb. Finally, the `across PP' and `across RC' structures test the effect of adding a noun (and verb in the latter structure) between the main subject and the main verb. In the `across RC' and `within RC' structures, we measure effects both with and without the complementizer \emph{that}.\footnote{A comparison of total and indirect effects when including or excluding the complementizer may be found in Appendix~\ref{app:complementizer}.}

In each of these constructions, we define a correct and an incorrect continuation. Here, we focus on the third-person singular/plural distinction.

\subsection{Models}
We focus primarily on GPT-2 \citep{gpt2}, an autoregressive Transformer-based \citep{transformer} English LM. We use several GPT-2 sizes, including DistilGPT-2 \citep{sanh2020distilbert}, a very small distilled version. Table \ref{tab:sizes} gives model details for the different sizes of GPT-2.

\begin{table}
    \centering
    \resizebox{0.9\linewidth}{!}{
    \begin{tabular}{lrrr}
    \toprule
         Size & Layers & Embedding size & Heads \\ 
         \midrule
         Distil & 6 & 768 & 12 \\
         Small & 12 & 768 & 12 \\
         Medium & 24 & 1024 & 16 \\
         Large & 36 & 1280 & 20 \\
         XL & 48 & 1600 & 25
         \\ \bottomrule
    \end{tabular}}
    \caption{GPT-2 sizes used in this study. ``Embedding size'' and ``heads'' refer to the number of neurons and attention heads per layer, respectively.}
    \label{tab:sizes}
\end{table}

To investigate how differences in training across Transformer-based architectures manifest themselves in syntactic agreement mechanisms, we also investigate Transformer-XL \citep{transformerxl} and XLNet \citep{xlnet}. Transformer-XL is an autoregressive English LM whose training objective is similar conceptually to GPT-2's; however, it has a much longer effective context. XLNet is an English LM which proceeds through various word order permutations of the input tokens during training, and which uses a distinct attention masking mechanism as well; during testing, it proceeds autoregressively through the input similar to the other two models.

\section{Total Effect: How strongly do models prefer correct forms?}
We use the relative probabilities of the correct and incorrect tokens as a measure of the preference of a model (parameterized by $\theta$) for the correct inflection of a verb $v\in\mathbf{v}$ given prompt $u\in\mathbf{u}$ with number feature $\textit{sg}$:
\begin{align}
    y(u_\textit{sg}, v) = \frac{p_\theta(v_\textit{pl}\mid u_\textit{sg})}{p_\theta(v_\textit{sg}\mid u_\textit{sg})}
    \label{eq:y}
\end{align}
where $y < 1$ indicates a preference for the correct inflection, and $y > 1$ indicates a preference for the incorrect inflection.\footnote{We arbitrarily choose to start with \emph{sg}; we can swap \emph{sg} and \emph{pl} in Eq.~\ref{eq:y} without loss of generality since we do not directly observe $y$. This is clarified after Eq.~\ref{eq:te}.}

To obtain counterfactual inputs, we now define a class of interventions $\mathbf{x}$ that modify the prompts in $\mathbf{u}$ in a systematic way. As we are concerned with the ability of models to choose correct inflections despite the presence of distractors and attractors, we define the intervention \texttt{swap-number}, which replaces the target subject with the same lexeme of the opposite number inflection (e.g., change ``author'' to ``authors'' or vice versa).
We also define the \texttt{null} intervention, which leaves $u$ as-is (as in \citealt{vig2020causal}).

Now we define $y_x(u,v)$, which is the value of $y$ under intervention $x$ on prompt $u$. Because the intervention \texttt{swap-number} entails swapping the subject for a noun of the opposite number, we now expect $y > 1$ in Equation~\ref{eq:y} if the model prefers the grammatically correct form, since the verb that was originally the correct inflection is now incorrect and vice versa. Note that under this definition, $y_\texttt{swap-number}(u_\textit{sg},v) = 1/y_\texttt{null}(u_\textit{pl},v)$.

\begin{figure}
    \centering
    \includegraphics[width=\linewidth]{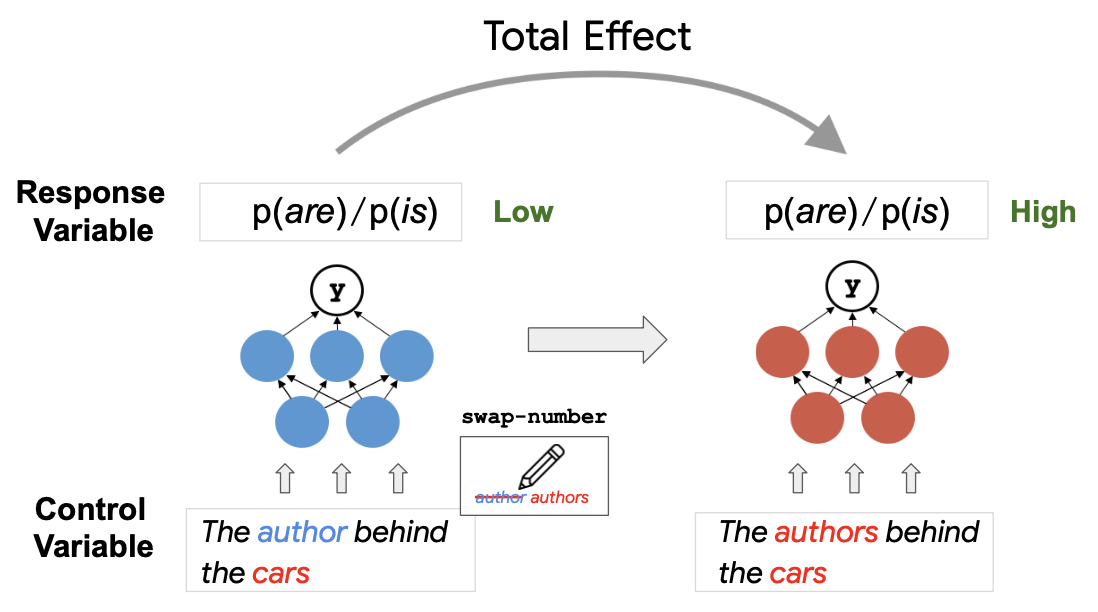}
    \caption{Total effects are measured by performing an intervention on the prompt (here, changing the grammatical number of the main subject), and measuring the relative change in the response variable (the ratio of probabilities of the originally incorrect verb form over the originally correct verb form).}
    \label{fig:te-illustration}
\end{figure}

The total effect (TE) for the intervention $\texttt{swap-number}$ (illustrated in Figure \ref{fig:te-illustration}) is the relative change between the probability ratio $y$ under the \texttt{swap-number} intervention and the ratio under the \texttt{null} intervention:
\begin{align}
    \begin{split}
    \text{TE}(\texttt{swap-number, null};y,u,v) &= \\
    \frac{y_{\texttt{swap-number}}(u_\textit{sg},v) - y_{\texttt{null}}(u_\textit{sg},v)}{y_{\texttt{null}}(u_\textit{sg},v)} &= \\
    y_{\texttt{swap-number}}(u_\textit{sg},v)/y_{\texttt{null}}(u_\textit{sg},v) - 1 &= \\
    1/(y_\texttt{null}(u_\textit{sg},v)\cdot y_\texttt{null}(u_\textit{pl},v)) -1
    \end{split}
    \label{eq:te}
\end{align}

We interpret this quantity as the overall preference of a model for the correct inflection of $v$ in context $u$. Observe that this definition remains the same when \textit{sg} and \textit{pl} are swapped in Equation~\ref{eq:te}, therefore we do not specify whether $u$ is plural or singular in $\text{TE}(\texttt{swap-number, null};y,u,v)$.
    
We are interested in the average total effect across prompts and verbs:
\begin{align}
    \begin{split}
    \overline{\text{TE}}(\texttt{swap-number, null};y) &=\\ \mathbb{E}_{u,v}\left[ \frac{y_\texttt{swap-number}(u,v)}{y_\texttt{null}(u,v)} - 1 \right]
    \end{split}
\end{align}

We calculate the average total effect for each syntactic construction for different sizes of GPT-2 and consider other models later on. As a control, we also calculate total effects for models with random weights. Unlike in \citet{dupouxlinzen16}, we do not measure accuracies by checking whether one probability is higher than another. Rather, the total effect quantifies the \emph{margin} between the probabilities of correct and incorrect continuations with some intervention.

Because larger models tend to exhibit correct subject-verb agreement more often than smaller ones~\citep{hu2020systematic,martylinzen19}, we hypothesize that larger models will generally have larger TEs for the same structure (i.e., we predict that higher accuracy is indicative of larger margin between probabilities).

\subsection{Results}

\begin{figure*}[ht!]
    \includegraphics[width=0.95\linewidth]{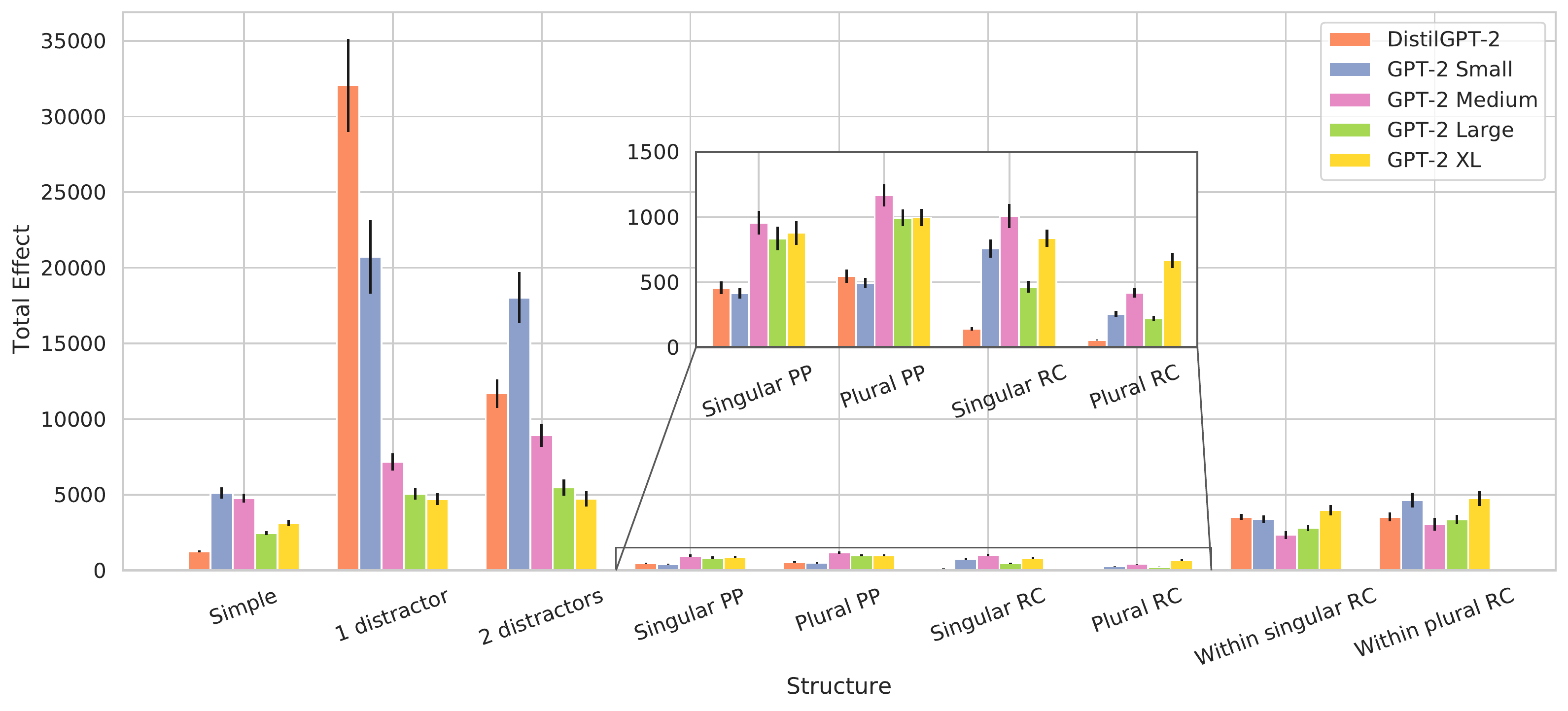}
    \caption{Total effects for each structure by model size for GPT-2. Adverbial distractors increase total effects, while attractor phrases decrease them.}
    \label{fig:total-effects}
\end{figure*}

Figure~\ref{fig:total-effects} presents total effects by structure for various sizes of GPT-2. For models with random weights, TEs are always near-zero, and as such are not shown in the figure.

In `simple agreement' and `within RC', where there is no separation of subject and verb, TEs vary between 1,000 and 5,000, depending on model size. This is far higher than the TEs below 250 reported for gender bias in \citet{vig2020causal}, which is to be expected: GPT-2's training objective explicitly optimizes for predicting (ideally grammatically correct) tokens given a context. Unlike \citet{vig2020causal}, we do not observe larger TEs for larger models.

\textbf{Adverbial distractors increase total effects.}
TEs are even higher for structures where distractors are present, with DistilGPT-2 and GPT-2 Small attaining the highest TEs in such contexts.  This is surprising, as one might expect subject-verb agreement accuracy to decline as the distance between the subject and the verb increases. We suspect that adverbs are acting as cues that a verb will soon appear, thus increasing the probability of both the correct and incorrect verb, but increasing that of the correct verb more (for similar findings in human sentence processing, see~\cite{vasishth2006argument}). Additional analysis supports this hypothesis; see Appendix~\ref{app:adverbs}.

\textbf{Attractors decrease total effects.} When PPs or RCs separate the subject and verb, TEs decrease. The number of the attractor does not significantly change TEs across PPs, but does have a more notable effect across RCs: GPT-2 is more certain of its choices across singular RCs than across plural RCs, as evidenced by higher TEs for the former. Notably, GPT-2 Medium tends to achieve the highest TEs in attractor structures, except in the `across plural RC' structure.

\section{Grammaticality Margin: Is agreement easier for singular or plural subjects?}\label{sec:gram}

Total effect measures the effect of swapping the number of the subject, but does not distinguish the case where the original subject (before swapping) was singular from the case where it was plural. To investigate the effect of the original subject number on the model's preference for the correct (or incorrect) inflection, we define the metric \textit{grammaticality margin} (referred to hereafter as \textit{grammaticality}) as the reciprocal of $y$ given prompt $u$ with a specific number feature \textit{sg} or \textit{pl}:
\begin{align}
\begin{split}
    G(u_\textit{sg}, v) &= 1/y(u_\textit{sg},v) \\ 
    G(u_\textit{pl}, v) &= 1/y(u_\textit{pl},v) \\
\end{split}
\end{align}
Recalling the definition of $y$, this measure is the probability ratio between the model correctly and incorrectly resolving subject-verb agreement. We define $G$ as the reciprocal of $y$ so that when the model has a high preference for the correct inflection over the incorrect inflection, $G$ is large.

Differences in grammaticality values for plural and singular subjects can indicate systematic biases toward a certain grammatical number. We expect this quantity to be lower if there is an attractor of a different number from the subject, whereas we expect it to increase if the attractor is of the same number as the subject.

\subsection{Results}
\begin{figure}
    \includegraphics[width=\linewidth]{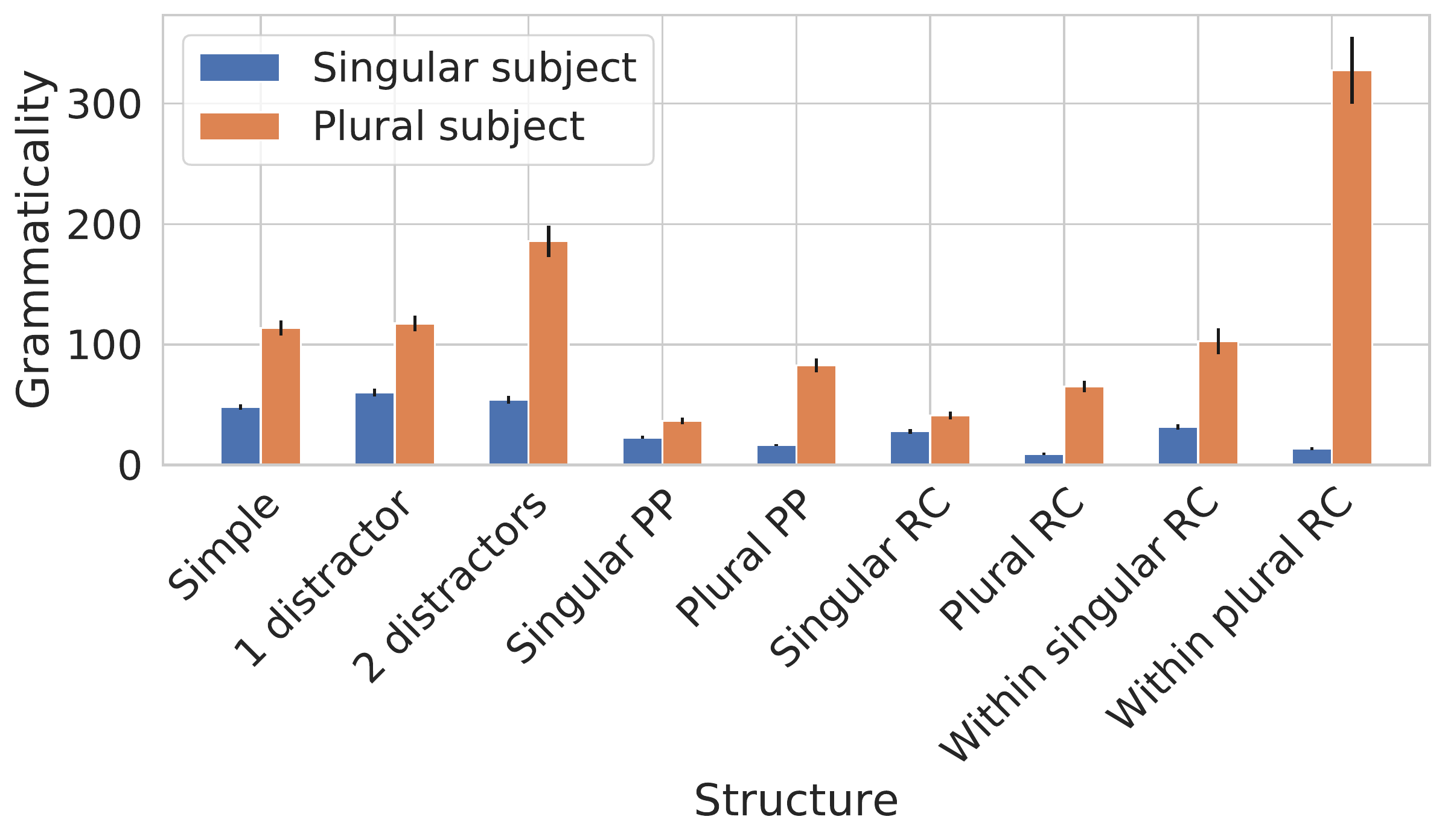}
    \caption{Grammaticality for each structure for GPT-2 Medium. The subject number (indicated by bar color) refers to the grammatical number of the subject with which the target verb agrees; the number in the structure name refers to the grammatical number of the attractor (in structures where attractors are present).}
    \label{fig:grammaticality}
\end{figure}

Figure~\ref{fig:grammaticality} presents grammaticality values separately for singular and plural subjects, as well as singular and plural attractors when applicable. While we expect higher grammaticality values when the subject number matches the attractor number, we instead observe that \textbf{plural subjects always have higher grammaticality values regardless of the structure or attractor number.} In other words, it is always easier for GPT-2 to form agreement dependencies between verbs and \emph{plural} subjects than singular subjects. This may be due to plural verbs being encoded as ``defaults'' in GPT-2, as was found for LSTM LMs in \citet{jumelet2019defaults}. This would make intuitive sense, because singular third person verbs are marked in English present-tense.

\textbf{Attractors that separate subjects and verbs decrease grammaticality, regardless of plurality.} The same is not true of distractors: placing adverbs between the subject and verb tends to have little effect, even though the `across two distractors' structure places the same token distance between subject and verb as `across a PP'. This means that distance between subject and verb is less important than the type of the structure separating them.

As expected, when holding the subject number constant (i.e., looking only at blue bars or only at orange bars in Figure~\ref{fig:grammaticality}), grammaticality values are higher when the attractor has the same number as the subject.

\textbf{Attractors that precede subjects have number-dependent impacts on grammaticality.} In the `within singular RC' structure, grammaticality is only slightly reduced for both singular and plural subjects compared to the `simple agreement' structure. However, `within plural RC' has a polarizing effect: grammaticality is greatly reduced for singular subjects, but greatly increased for plural subjects. This is the only attractor structure with higher grammaticality than the simple case.

\section{Natural Indirect Effect: Which components mediate syntactic agreement?}
\begin{figure}
    \centering
    \includegraphics[width=\linewidth]{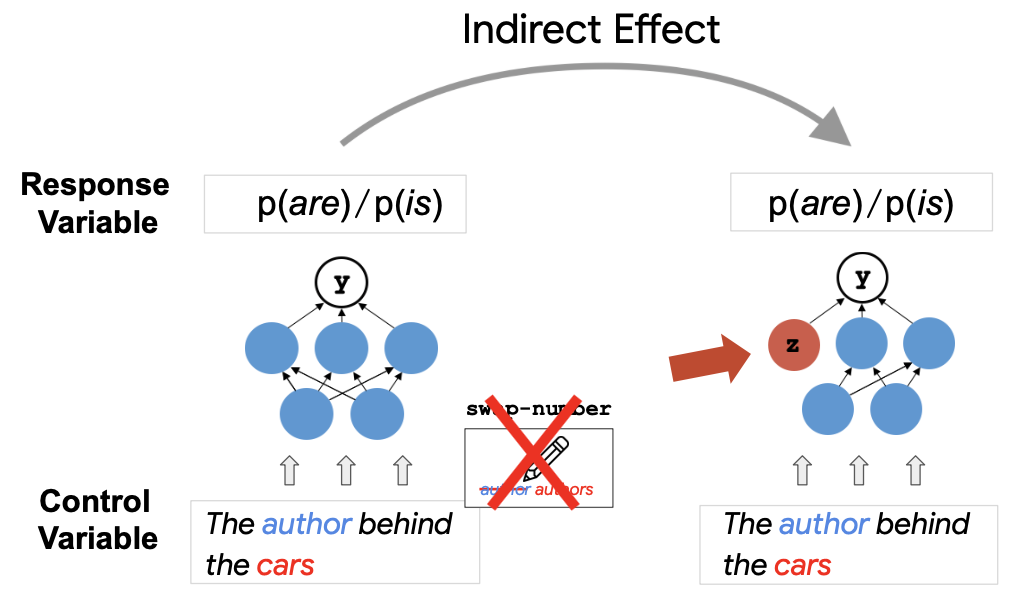}
    \caption{Indirect effects are measured by setting an individual neuron to the value it would have taken had the intervention occurred, then measuring the relative change in the response variable.}
    \label{fig:ie-illustration}
\end{figure}

The natural indirect effect (NIE), illustrated in Figure~\ref{fig:ie-illustration}, is the relative change in the ratio $y$ when the prompt $u$ is not changed, but a model component $\mathbf{z}$ (e.g., a neuron) is set to the value it \emph{would have taken} if the intervention had occurred. 
\begin{align}
    \begin{split}
    \overline{\text{NIE}}(\texttt{swap-number, null};y,\mathbf{z}) = \\
    \mathbb{E}_{u,v}\left[ \frac{y_{\texttt{null},\mathbf{z}_{\texttt{swap-number}}(u,v)}(u,v)}{y_\texttt{null}(u,v)} - 1 \right]
    \end{split}
\end{align}
This allows us to evaluate the contribution of specific parts or regions of a model to the syntactic preferences we observe. More specifically, we can measure to what extent the total effect of swapping the subject on inflection preferences can be attributed to specific neurons. 
    
\begin{figure*}[ht]
    \centering
    \begin{minipage}{0.32\linewidth}
    \includegraphics[width=\linewidth]{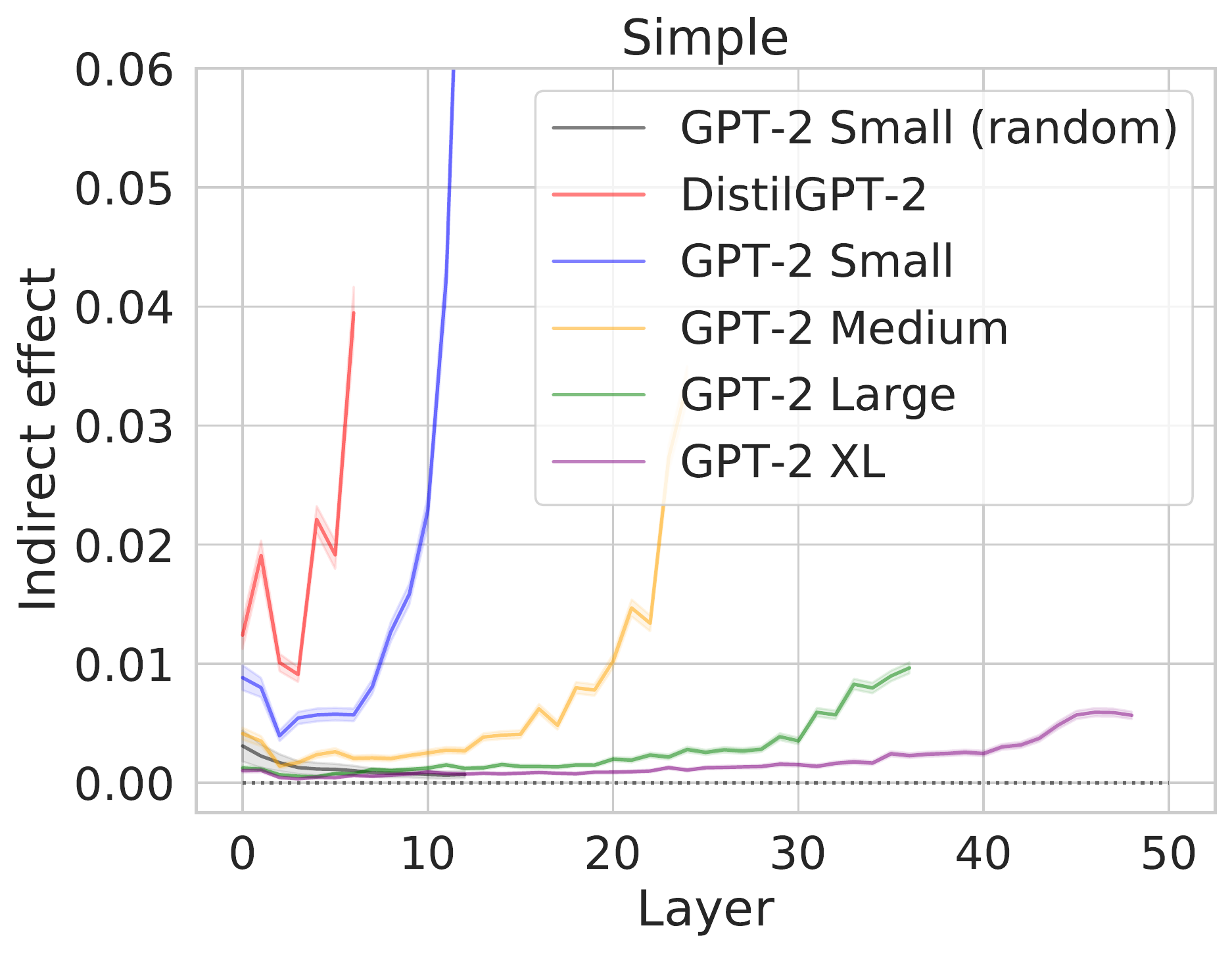}
    \end{minipage}
    \begin{minipage}{0.32\linewidth}
    \includegraphics[width=\linewidth]{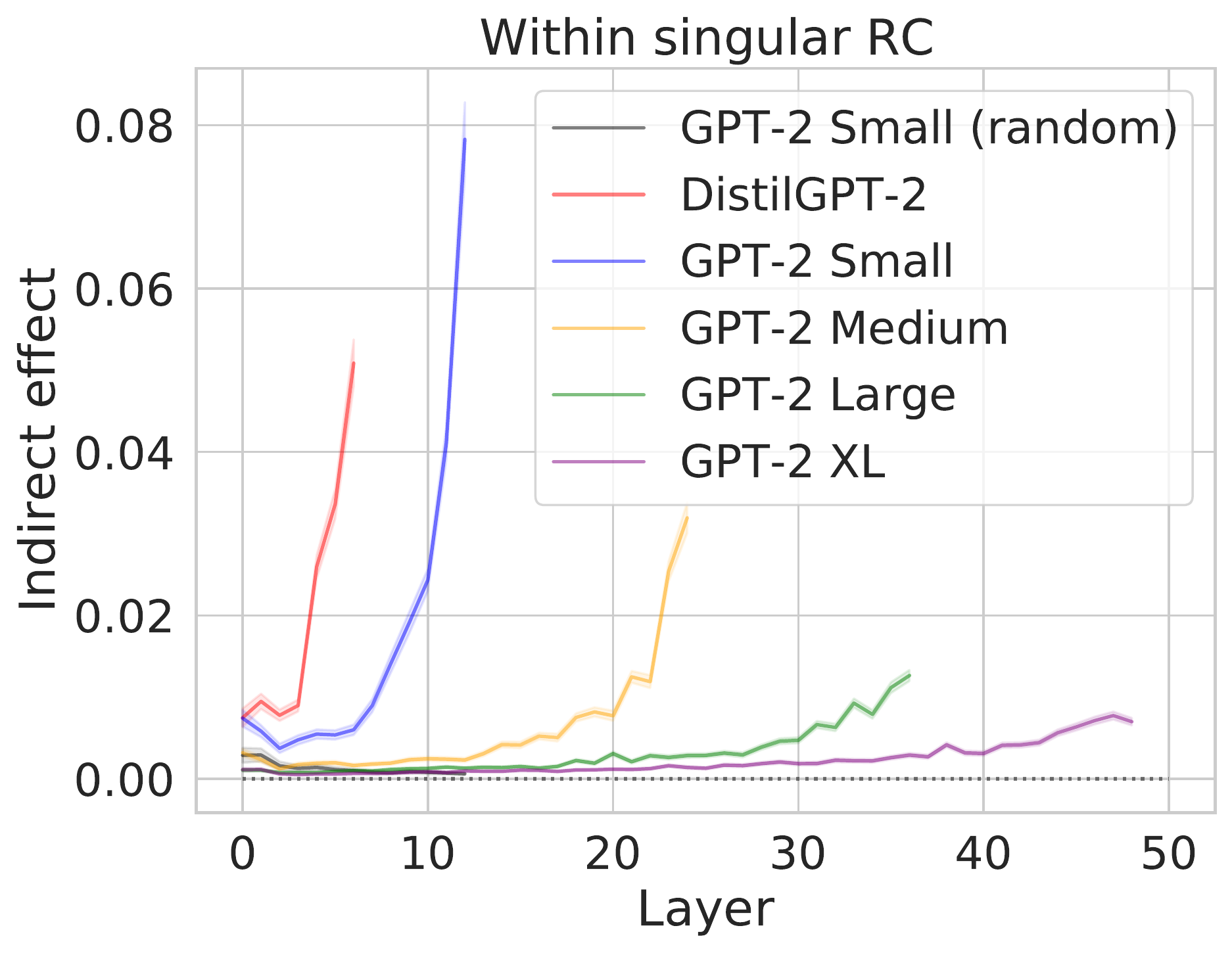}
    \end{minipage}
    \begin{minipage}{0.32\linewidth}
    \includegraphics[width=\linewidth]{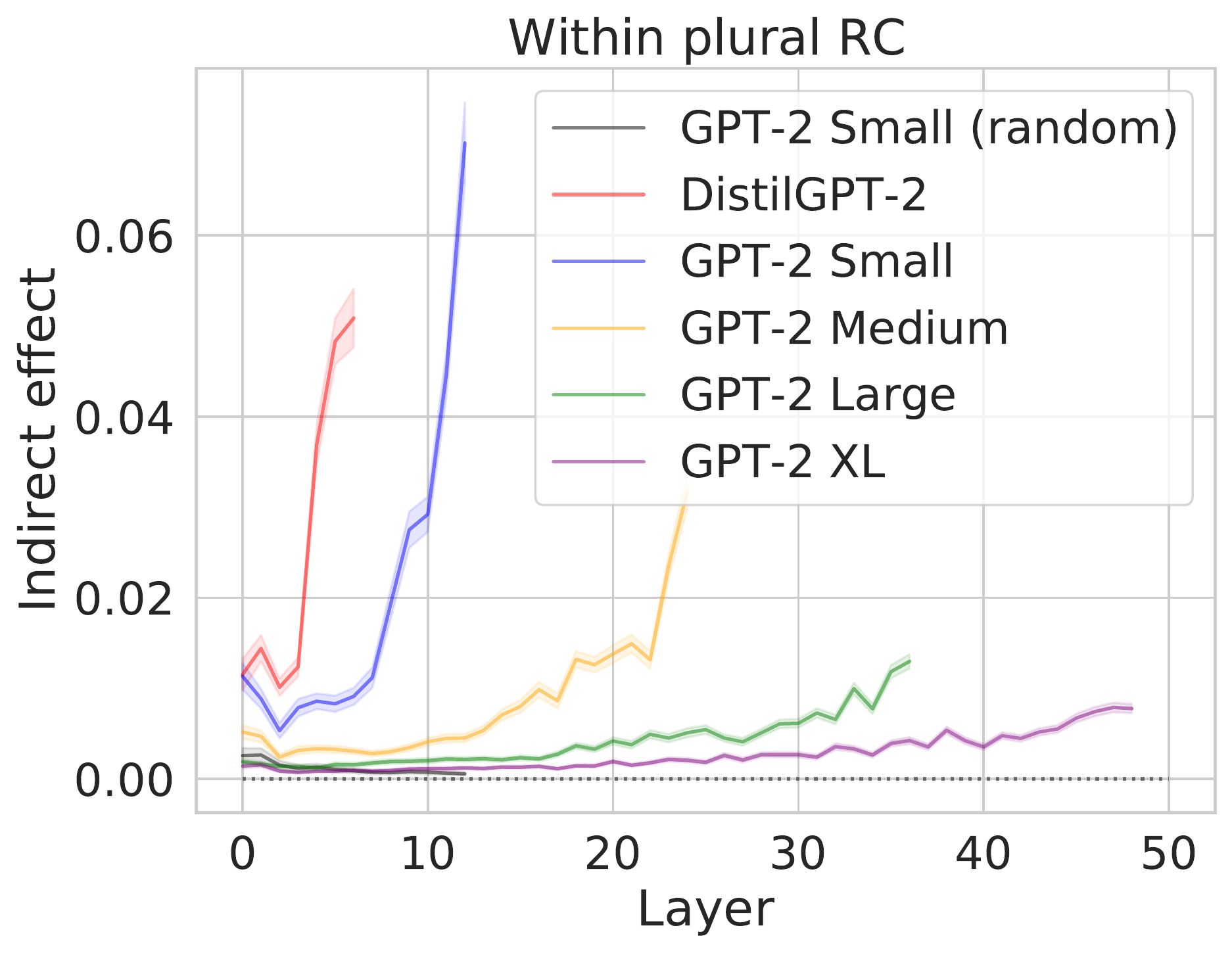}
    \end{minipage}
    \begin{minipage}{0.32\linewidth}
    \includegraphics[width=\linewidth]{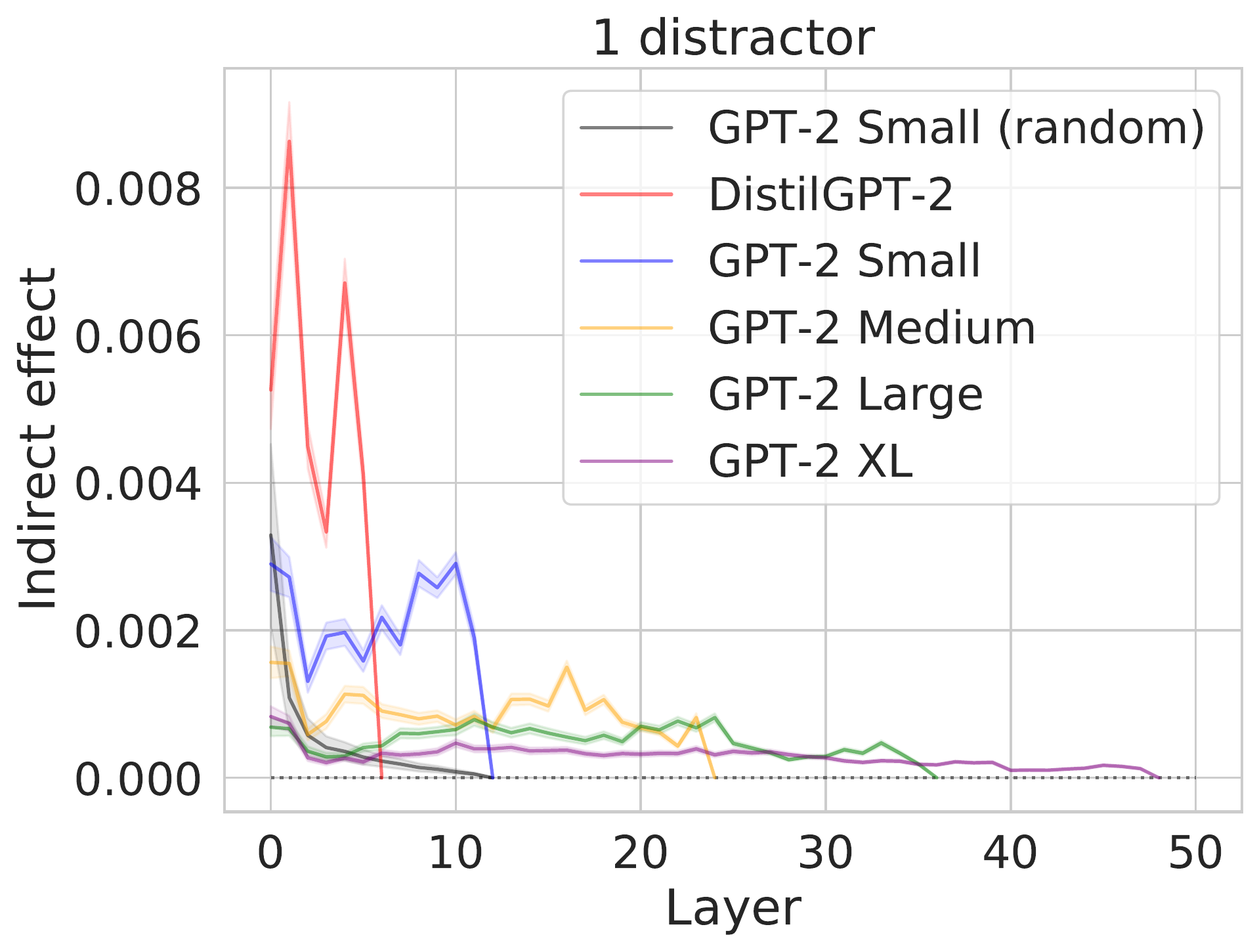}
    \end{minipage}
    \begin{minipage}{0.32\linewidth}
    \includegraphics[width=\linewidth]{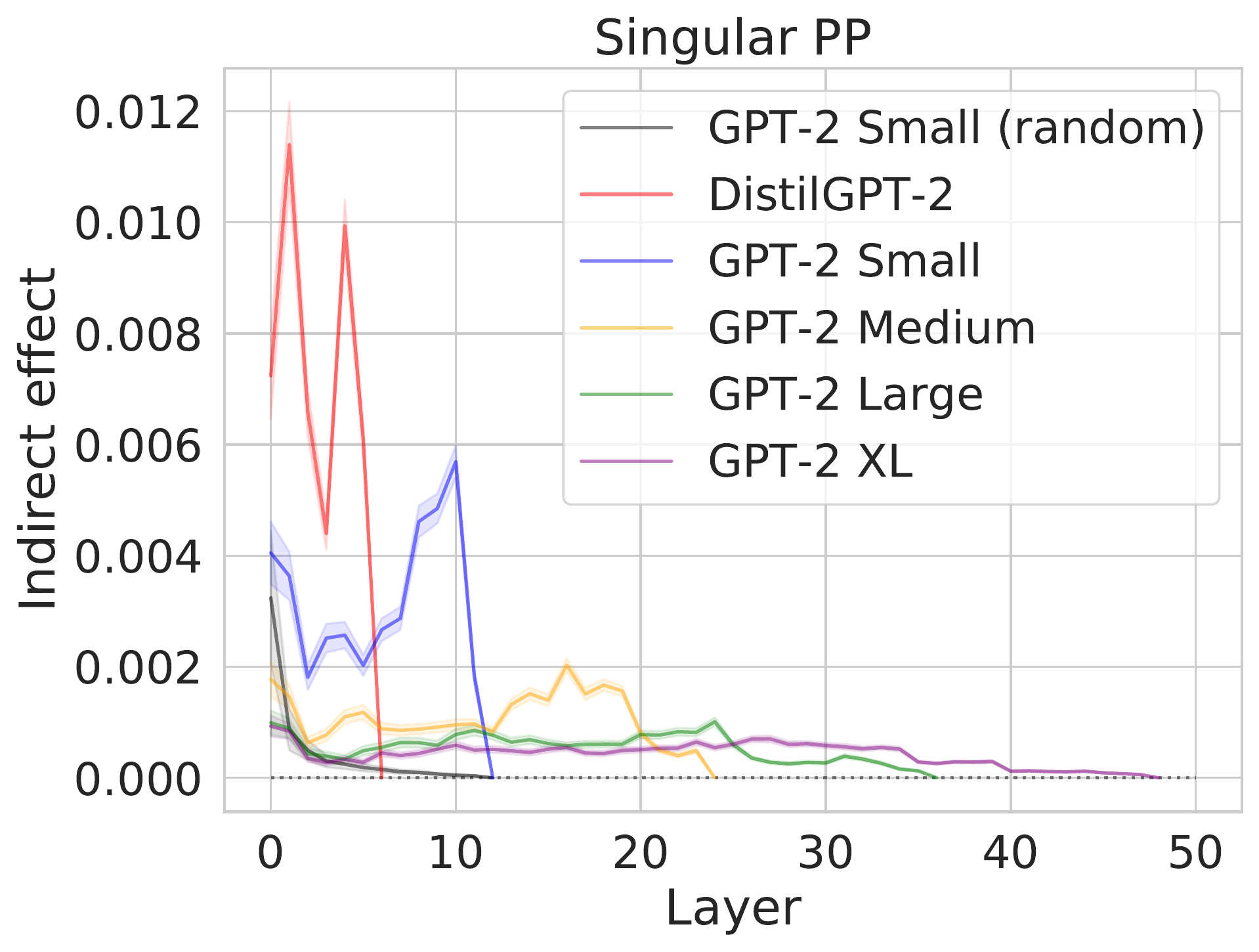}
    \end{minipage}
    \begin{minipage}{0.32\linewidth}
    \includegraphics[width=\linewidth]{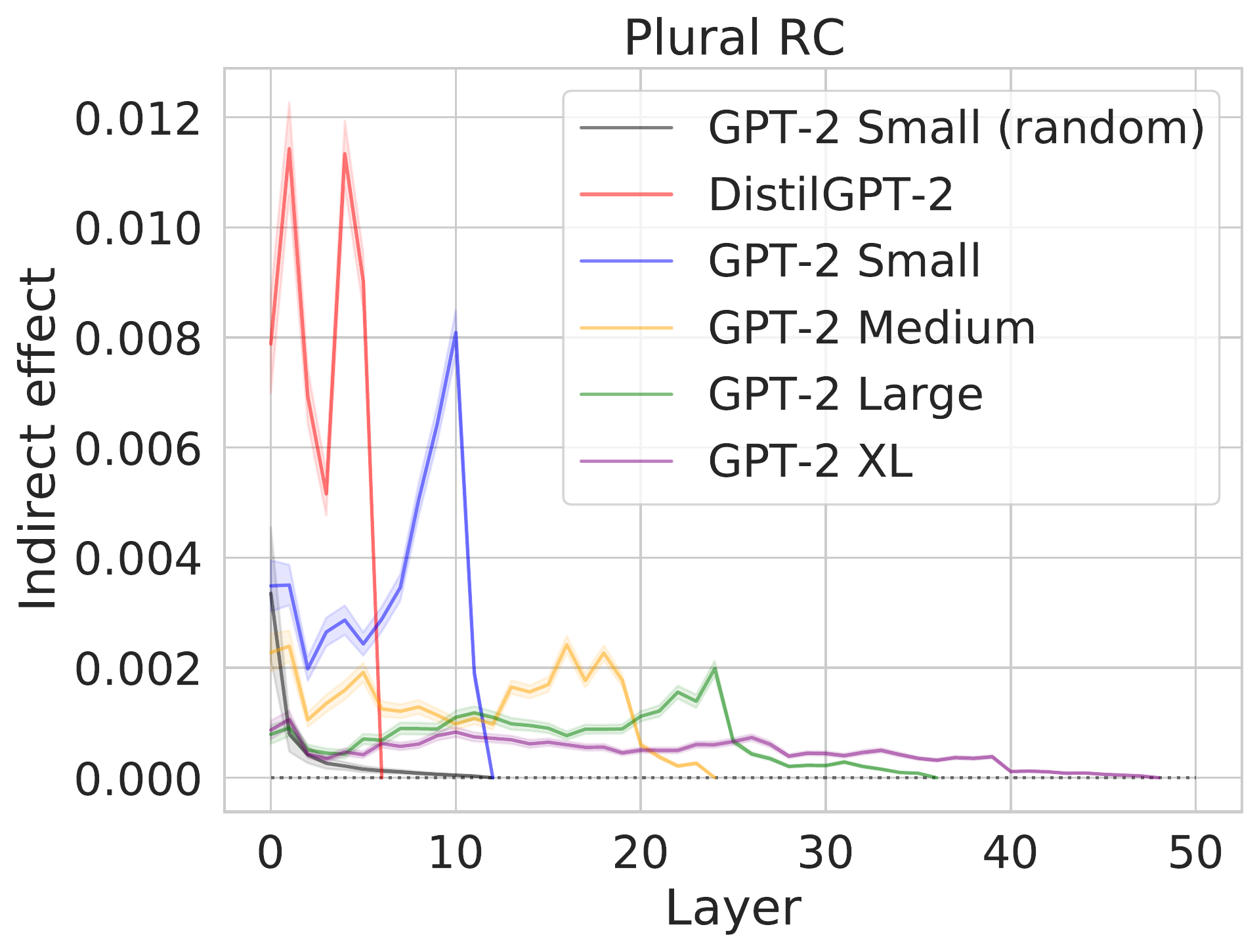}
    \end{minipage}
    \caption{Natural indirect effects of the top 5\% of neurons in each layer of various GPT-2 sizes. Each figure focuses on a single structure and compares across GPT-2 sizes.}
    \label{fig:neuron-sizes-top5}
\end{figure*}

\begin{figure}
    \centering
    \includegraphics[width=0.8\columnwidth]{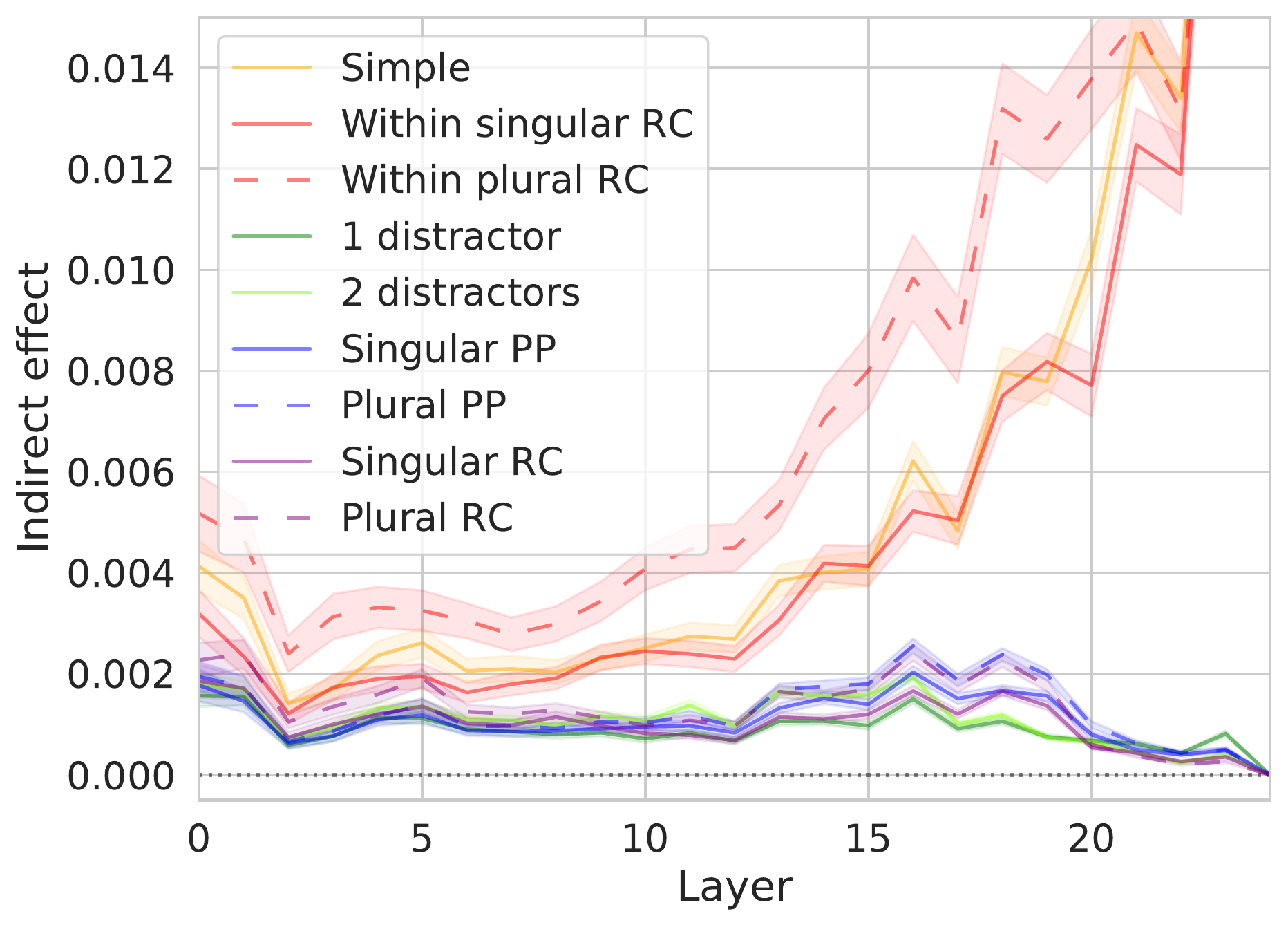}
    \caption{Natural indirect effects of the top 5\% of neurons in each layer of GPT-2 Medium.}
    \label{fig:neuron-structures-top5}
\end{figure}

Here, we independently analyze the individual neuron NIEs for GPT-2, Transformer-XL, and \mbox{XLNet} (future work could also investigate intervening on sets of neurons simultaneously). We also attempt to analyze attention heads for GPT-2, though we find that they do not present consistent interpretable results with the \texttt{swap-number} intervention (see Appendix~\ref{app:attn}). This is consistent with the findings of \citet{htut2019attn} who do not find a straightforward connection between attention weights and the model's syntactic behavior. 

Based on the findings of prior probing work on dependency parsing~\citep{hewitt2019structural}, we hypothesize that NIEs will peak in the upper-middle layers for all models. Because XLNet is exposed to all word order permutations of its input sentences during training, we hypothesize that it will display similar indirect effect results across syntactic structures. Conversely, GPT-2 and Transformer-XL always process input left-to-right, so we expect that for these two models, differing syntactic structures will yield unique indirect effect results.

\subsection{Results}\label{sec:nie_results}

For each model and structure, we select the 5\% of neurons with the highest NIE in each layer; Figure~\ref{fig:neuron-sizes-top5} compares NIEs across model sizes, and Figure~\ref{fig:neuron-structures-top5} compares NIEs across structures for GPT-2 Medium.\footnote{We also produced figures using \emph{all} neurons. When doing so, the contour of the graph across layers did not change, but the magnitudes were lower since we average over more neurons.} We observe two distinct layer-wise contour patterns. In structures where the target verb directly follows the subject (`simple agreement' and `within RC', the top 3 plots in Figure~\ref{fig:neuron-sizes-top5}), NIEs continually increase in higher layers.

Conversely, for structures with subject-verb separation (`across one/two distractor(s)', `across PP', and `across RC', the bottom 3 figures in Figure~\ref{fig:neuron-sizes-top5}), NIEs peak at layer 0 and (more notably) in the upper-middle layers. This is in line with the probing results of \citet{hewitt2019structural} and \citet{tenney2019pipeline}, who find that the highest amount of syntactic information is encoded in the upper-middle layers. In the final layers of the model, the effect decreases sharply, reaching 0 in the uppermost layers. The peak NIE is lower here than for structures where there is no separation, perhaps indicating that syntactic agreement information is localized in fewer neurons when separation occurs.

Even a single token between subject and verb brings about this second indirect effect contour, indicating that \textbf{distance is a less important factor than the presence of \emph{any} separation in invoking this second syntactic agreement mechanism}. The distinct indirect effect contours for the adjacent and non-adjacent cases may indicate distinct subject-verb agreement mechanisms for short- and long-distance agreement, consistent with similar findings for LSTMs \citep{lakretz2019emergence}.

As a control, we repeated the experiment for GPT-2 with randomized weights. We find that for all structures, when weights are randomized, indirect effects peak at layer 0---albeit at values perhaps too small to be meaningful---and then remain close to 0 in higher layers. This indicates that the vast majority of the indirect effect observed for trained models is an outcome of learning from the training data rather than of the architecture.

For each structure, the maximum NIE per layer is always lower for larger models. Peaks in NIEs are also more distributed across layers for larger models. \textbf{This suggests that structural knowledge is concentrated in fewer neurons with stronger inflectional preferences in smaller models, and is more distributed across neurons in larger models.} Nonetheless, the overall contour of NIEs is similar across model sizes for a given structure, indicating that \textbf{mechanisms of agreement are similar across model sizes}.

\subsubsection{Comparing GPT-2 to Other Architectures}\label{sec:ie_architectures}
\begin{figure}[t]
    \centering
    \includegraphics[width=\columnwidth]{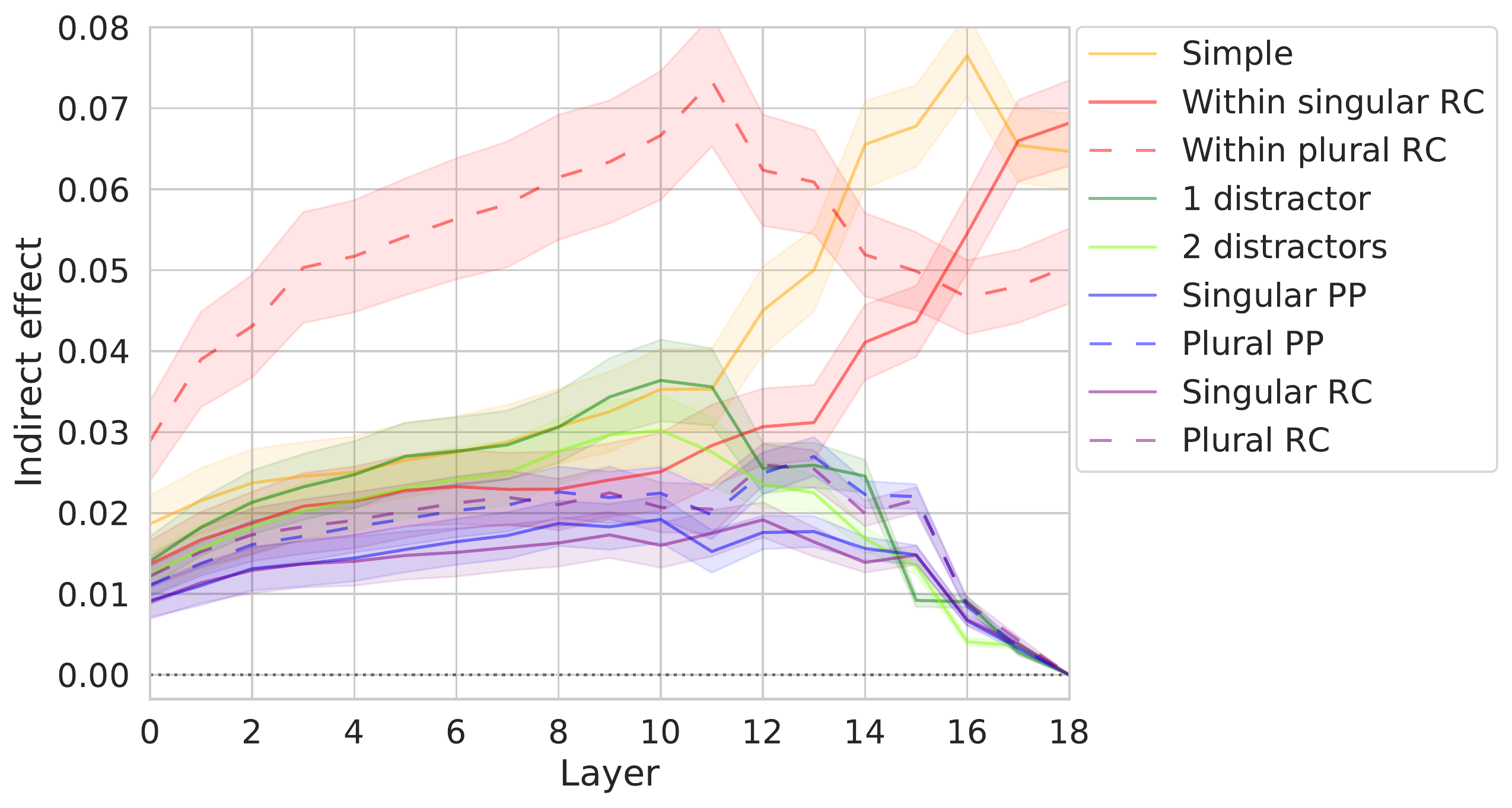}
    \caption{Natural indirect effects of the top 5\% of neurons in each layer of Transformer-XL.}
    \label{fig:txl_neuron}
\end{figure}

\begin{figure}[t]
    \centering
    \includegraphics[width=\columnwidth]{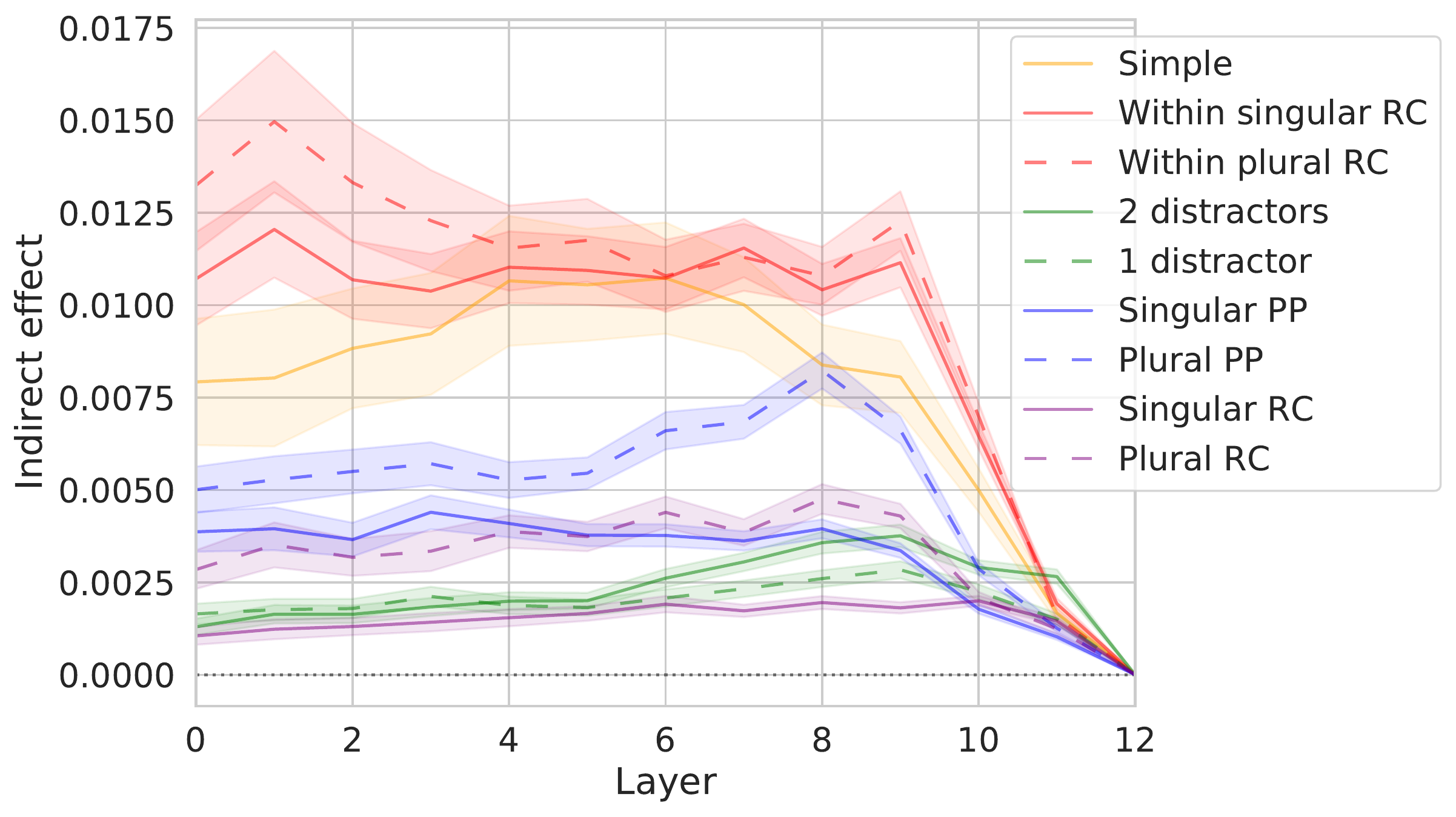}
    \caption{Natural indirect effects of the top 5\% of neurons in each layer of XLNet.}
    \label{fig:xlnet_neuron}
\end{figure}

We also investigate the neuron NIEs of Transformer-XL (Figure~\ref{fig:txl_neuron}) and XLNet (Figure~\ref{fig:xlnet_neuron}) to observe whether syntax is represented in a similar manner across models (for total effects across architectures, see Appendix~\ref{app:te_architectures}).

\textbf{Local and non-local agreement diverges in a similar way in GPT-2 and Transformer-XL.} The layer-wise contour is similar for `simple agreement' and `within RC' across the two architectures, and differs significantly from the cases where subject and verb are separated, which is again similar across architectures. This supports our hypothesis that GPT-2 and Transformer-XL encode syntax in a similar manner.

\textbf{Indirect effects in XLNet are different to those seen in GPT-2.} In XLNet, we do not observe the same dichotomous behavior between subject-verb adjacent and subject-verb non-adjacent structures; rather, the overall contours are all similar. All of the indirect effects approach 0 in the final layer. This resembles the contours from GPT-2 and Transformer-XL for structures where subject and verb are not adjacent. We conjecture that this pattern arises because XLNet observes many word order permutations of the same inputs during training; this acts as a form of regularization that prevents it from evolving bifurcating mechanisms for local and non-local dependencies.

While \citet{sinha2021masked} found that natural word order during pre-training matters little for downstream performance on tasks in benchmarks like GLUE \citep{glue}, they also found that randomizing word order greatly reduced model preferences for correct inflections in syntactic evaluation stimuli. This finding---coupled with the distinct word-order-dependent agreement mechanisms that we discover---suggests that models do make use of word order information, rather than just higher-order word collocation statistics.

\subsubsection{Neuron Overlap Across Structures}\label{subsec:overlap}
The layer-wise NIE contours in Section~\ref{sec:nie_results} show the NIE of the top neurons in each layer, but do not show which neurons make it into the top 5\%. To investigate whether the same neurons are implicated in subject-verb agreement across structures, we select the top 5\% of neurons per layer by NIE and calculate the proportion of these high-NIE neurons that overlap between each pair of structures.

\begin{figure*}[t]
    \centering
    \begin{minipage}{0.325\linewidth}
    \includegraphics[width=\linewidth]{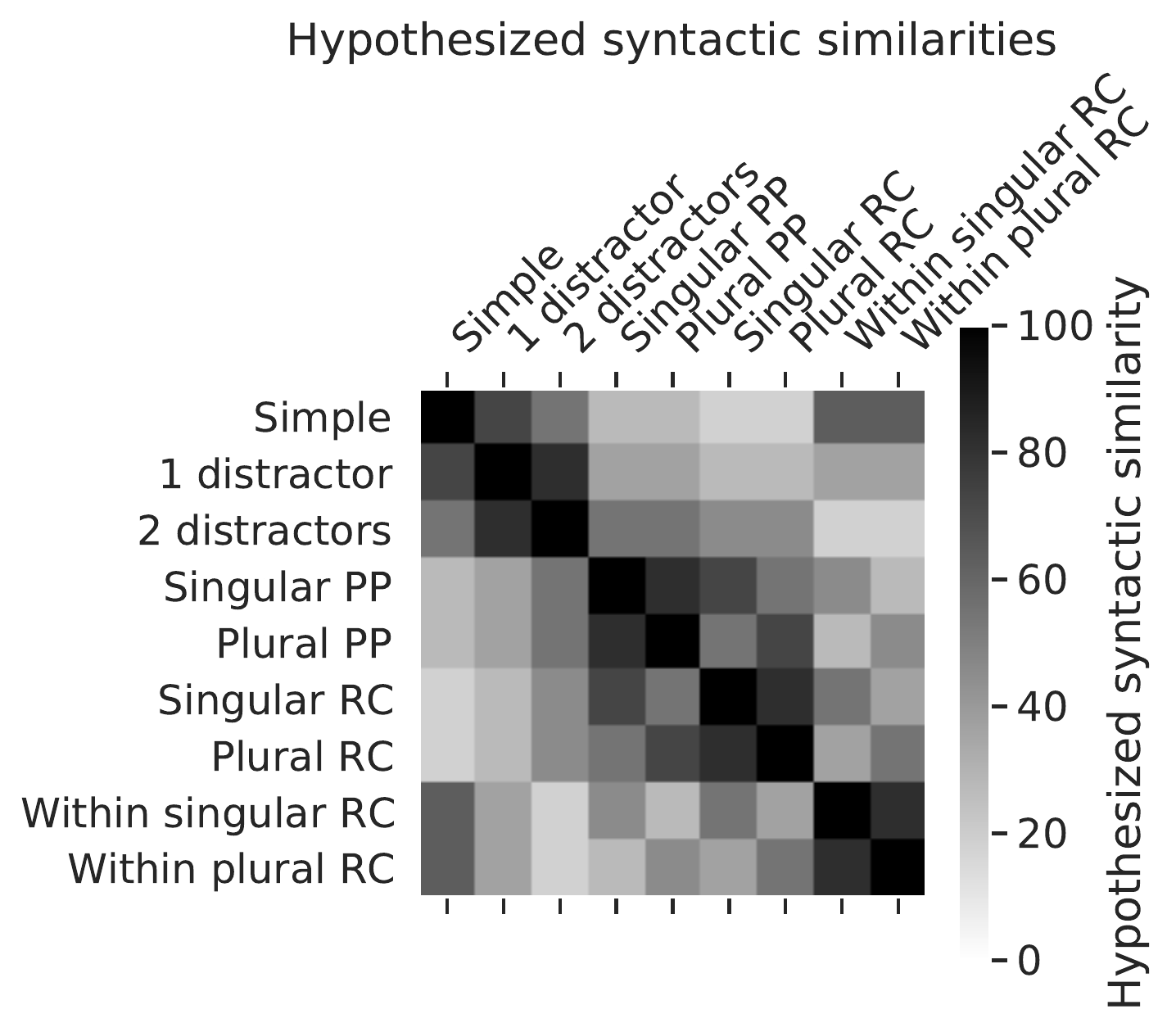}
    \end{minipage}
    \begin{minipage}{0.215\linewidth}
    \includegraphics[trim={4.87cm 0 0 0},clip,width=\linewidth]{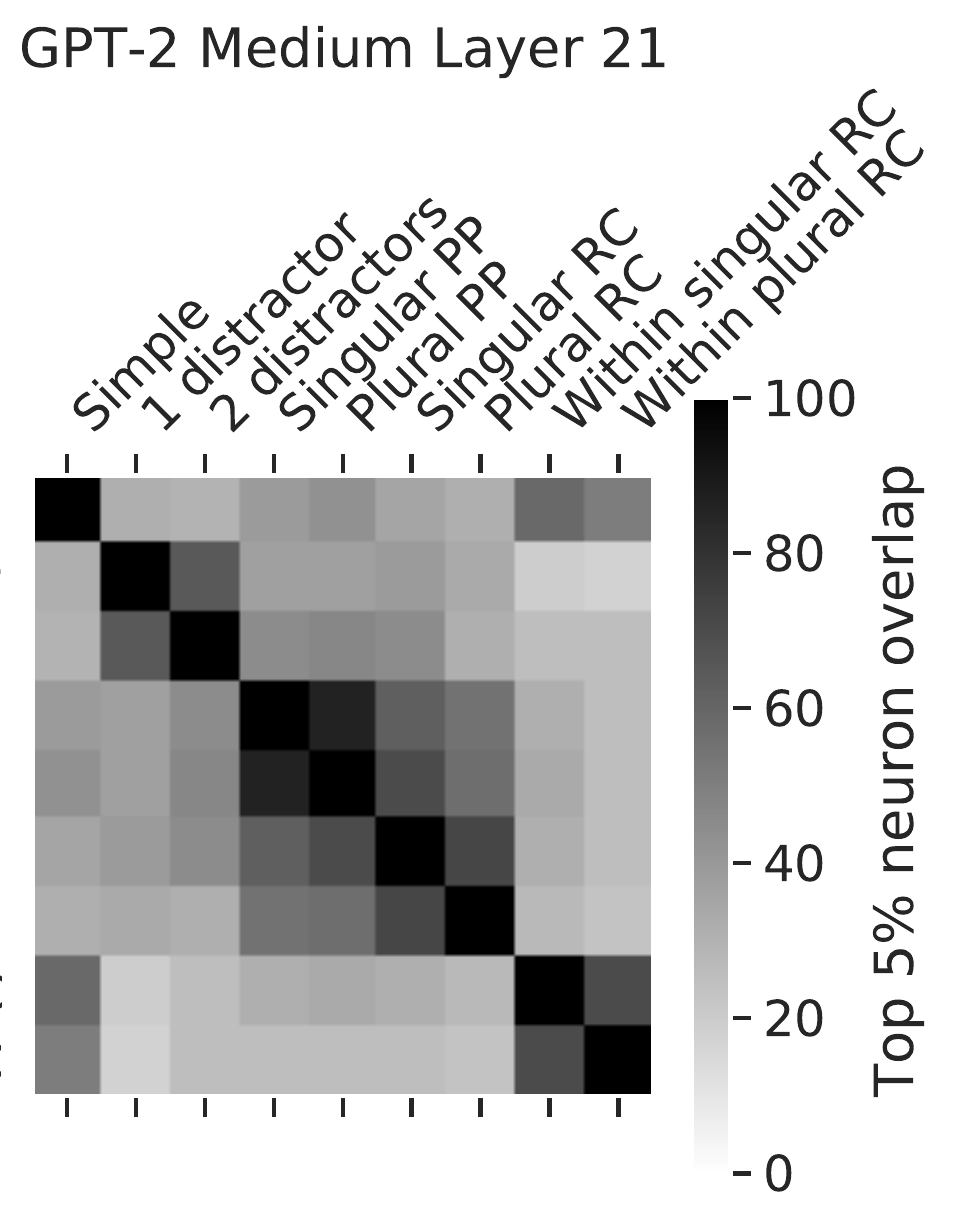}
    \end{minipage}
    \begin{minipage}{0.215\linewidth}
    \includegraphics[trim={4.87cm 0 0 0},clip,width=\linewidth]{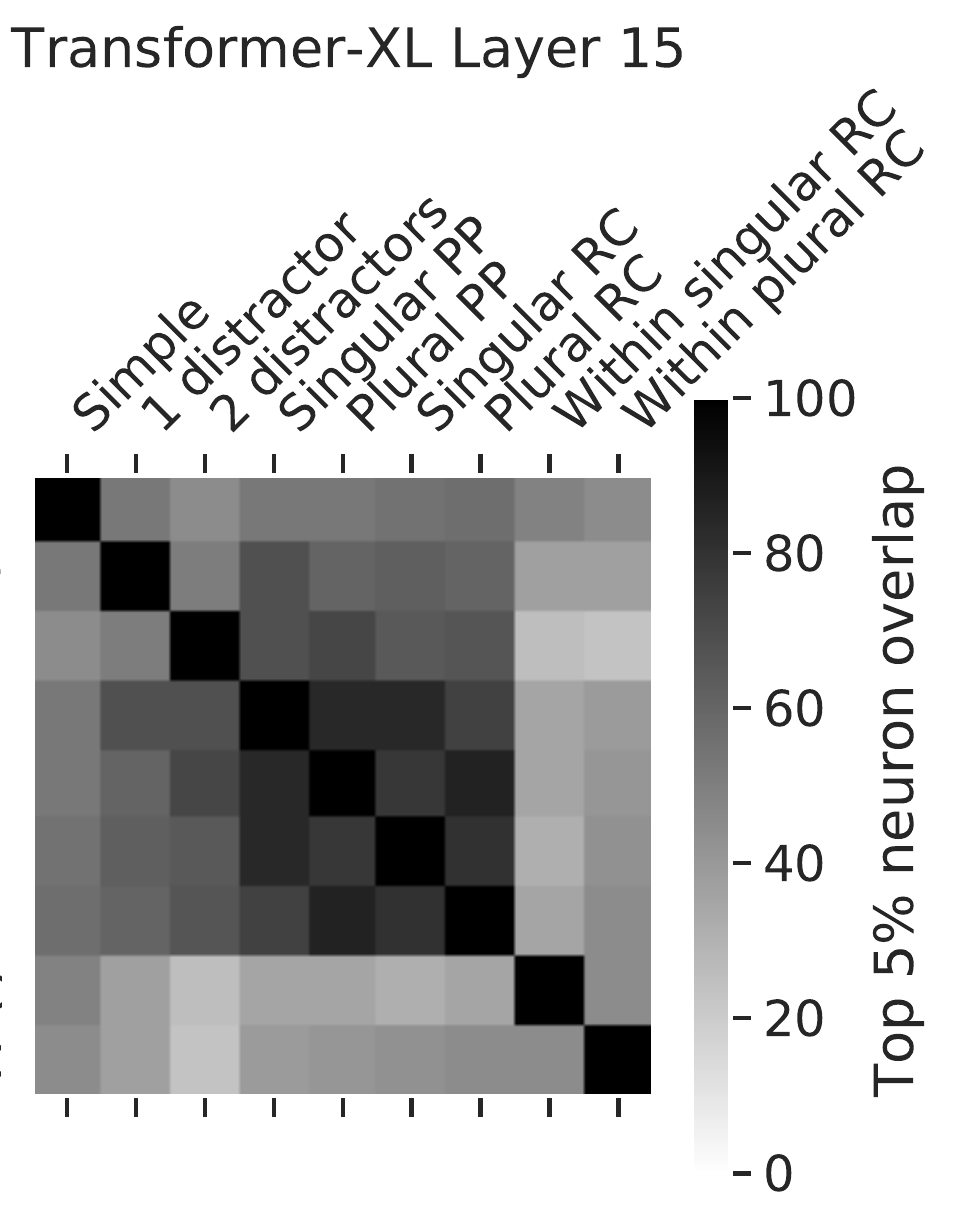}
    \end{minipage}
    \begin{minipage}{0.215\linewidth}
    \includegraphics[trim={4.87cm 0 0 0},clip,width=\linewidth]{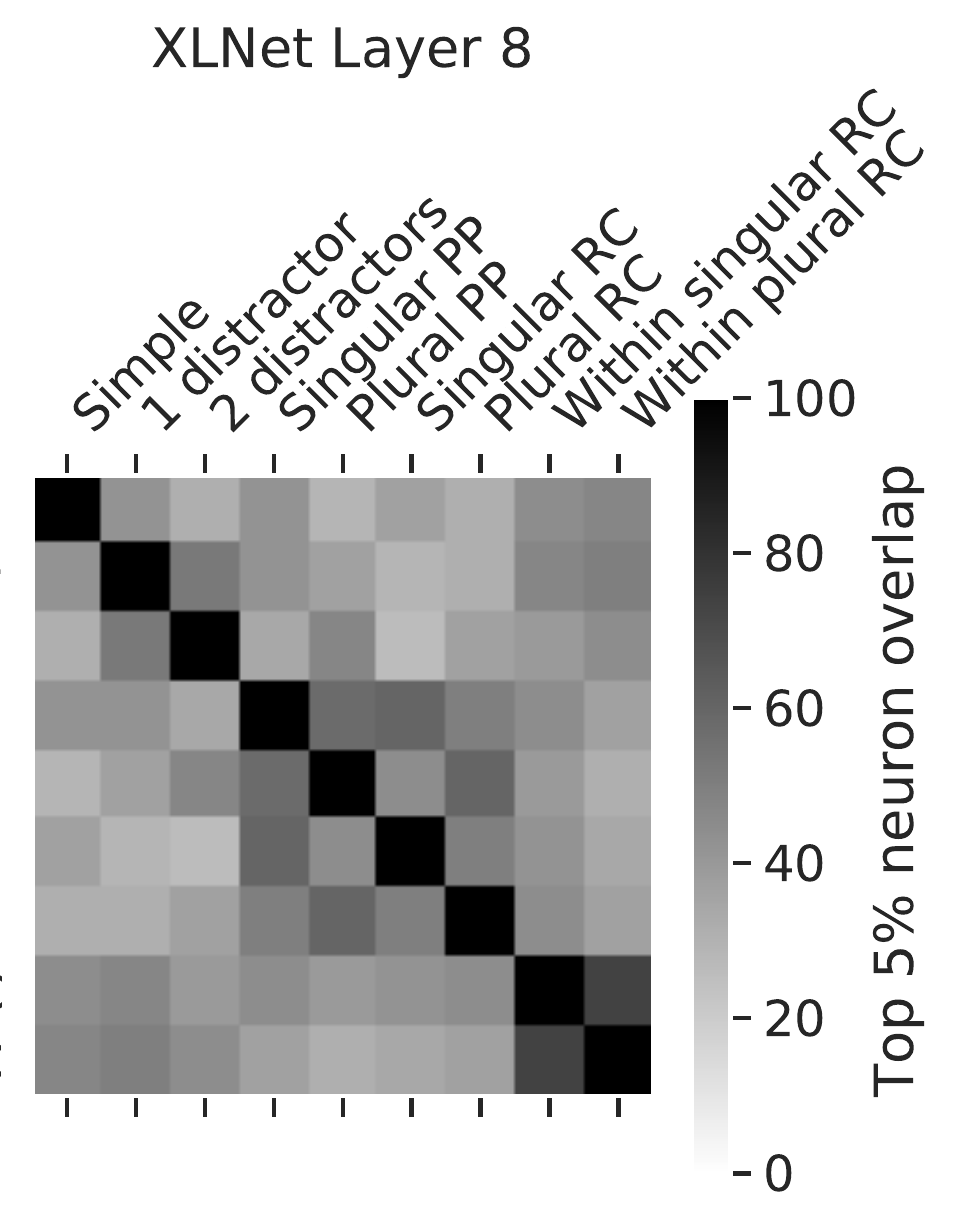}
    \end{minipage}
    \caption{Hypothesized syntactic similarity across structures (left), as well as the overlap of the top 5\% of neurons per-structure by indirect effect for GPT-2 (center-left), Transformer-XL (center-right), and XLNet (right); the layer displayed is the one that shows the highest similarity to the hypothesized (ground-truth) matrix.}
    \label{fig:neuron-overlap}
\end{figure*}

Does the extent of neuron sharing across structures correlate with human intuitions of syntactic similarity? To address this question, we compute hypothesized syntactic similarities between structures based on the following linguistic features: 
distance between subject and verb;
presence of adverbial distractors, a relative clause, prepositional phrase, and/or a noun attractor; 
and the number of the noun attractor when present. Appendix \ref{app:hypothesis_matrix} provides additional details on the calculation of ground-truth similarity.

To quantify the similarity of the hypothesis matrix and a neuron overlap matrix, we calculate the $\ell_1$ norm\footnote{Using the $\ell_2$ norm does not change which layer in each model has the lowest difference norm.} of the element-wise difference between the lower-left triangle of both matrices, as the matrices are symmetric. We exclude the diagonal.

For each model, we present neuron overlaps for the layer with the lowest difference norm to the hypothesis (Figure~\ref{fig:neuron-overlap}; for an analysis of layer-by-layer overlap change for GPT-2, see Appendix~\ref{app:overlap_by_layer}). The lowest difference norms are 443 (GPT-2), 510 (Transformer-XL), and 486 (XLNet). \mbox{GPT-2} Medium's overlap across structures at layer~21 (of~24) is visually similar to the hypothesis, indicating that \textbf{this layer in GPT-2 shares neurons for subject-verb agreement across structures in a way that aligns with human intuitions about syntactic similarity.} Interestingly, it learns to do this without receiving explicit syntactic supervision during training.

Layer 15 (of~18) of Transformer-XL displays similar trends to GPT-2, though the extent of overlap is higher across structures in general here. There is more significant overlap between the adverbial distractor structures and the structures that contain attractors. `Simple agreement' also has more overlap with structures containing attractors than `within RC', which is contrary to our hypothesis matrix. We also note that `across singular RC' has more overlap with `across PP' than `across plural RC' (and vice versa for `across plural RC'), indicating that \textbf{the number of the attractor is more salient to Transformer-XL than the structure of the phrase containing the attractor.}

Layer 8 (of 12) of XLNet gives rise to a noisier similarity matrix. There is slightly more overlap between structures across noun attractors, but the extent of overlap is smaller compared to other models. This suggests that more of the neurons are specialized to processing specific structures. However, the indirect effect findings for XLNet suggest a more unified mechanism for syntactic agreement across all structures; if this were the case, we would expect neuron overlap to be high, and for the extent of overlap to be similar across \emph{all} structures, rather than being higher between more similar structures. We observe the latter, but not the former. Regardless, both observations further support our hypothesis that XLNet uses different mechanisms to resolve number agreement than the other two architectures.

\section{Conclusions}

This study applied causal mediation analysis to discover and interpret the mechanisms behind syntactic agreement in pre-trained neural language models. Our results reveal the location and importance of various neurons within various models, and provide insights into the inner workings of these LMs.

For future work, we suggest intervening on groups of neurons and attention heads to see how these components work together, and extending the analysis to phenomena such as filler-gap dependencies and negative polarity items. Further work should also explore the impact of specific verbs on syntactic agreement mechanisms \cite{newman2021refining}. Lastly, we suggest examining examples where the model makes incorrect predictions to determine how models misuse the mechanisms from Section~\ref{sec:nie_results}.

\section*{Acknowledgements}
Y.B.\ was supported in part by the ISRAEL SCIENCE FOUNDATION (grant no.\ 448/20) and by an Azrieli Foundation Early Career Faculty Fellowship. A.M.\ was supported by a National Science Foundation Graduate Research Fellowship (grant no.\ 1746891).

% NOTE: When you upload the pdf file, softconf will mistakenly check whether (5a)+(5b) together, not (5a) alone, is within the page limit because impact statement is a recent addition to the pdf file and softconf has not updated its check tool to handle that. Thus, if sonfconf gives a warning saying that the pdf file is over the page limit, AND if that is caused by the paper having an impact statement on Page 5 (for short papers) or Page 9 (for long papers), you can ignore the warning. Just make sure that the main content and the Appendix do not go over the page limit.
\section*{Impact Statement}
% Our method limitations
In this paper, we apply causal mediation analysis in order to study the subject-verb agreement mechanisms in language models. While the focus of this work is on the analysis itself, our insights may influence the training strategies for new models. Specifically, our findings on the relationship between model size and syntactic agreement and the comparison of different model architectures may help researchers decide which model to use. In doing so, others may try to extrapolate our findings, which are limited to the domain of specific syntactic structures and subject-verb agreement in English language models, to other tasks and languages for which we cannot make these claims. The focus on English of this study additionally furthers the discrepancy compared to other languages which continue to be studied much less.

Moreover, we do not study mitigation mechanisms for our findings and thus do not know the consequences of modifying the training procedures of language models beyond the three  examples we studied. One concrete example for a case where our findings could have wider impact regards our finding that models have higher grammaticality for plural subjects. Others may find that this is undesired behavior and thus try to augment their training data to increase the number of subjects in singular form, which could have unanticipated consequences on model performance and mechanisms.

% causal mediation analysis limitations
Beyond the concrete findings in this paper, there are also broader considerations in the popularization of causal mediation analysis. Specifically, as pointed out by~\citet{vig2020causal}, it is a challenging problem to extend the effect measures beyond binary cases. While subject-verb agreement is by nature a binary problem, there are many others that benefit from a more nuanced view, specifically in topics related to fairness and bias. Thus, by popularizing an approach that is easier to apply in a binary case, we may have the unintended effect of complicating analyses conducted by others who want to follow our approach. As an active mitigation, we direct readers to the extended version of~\citet{DBLP:journals/corr/abs-2004-12265}, which discusses effect measures beyond the binary case.

\bibliography{cma_acl21}
\bibliographystyle{acl_natbib}

% APPENDIX
\appendix

\clearpage

\section{Attention Head Indirect Effects}\label{app:attn}
\begin{figure*}
    \centering
    \begin{minipage}{0.48\linewidth}
    \includegraphics[width=\linewidth]{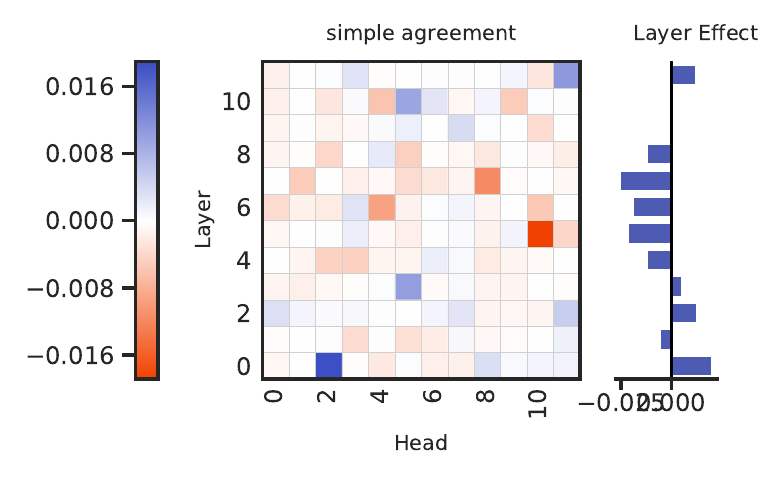}
    \end{minipage}
    \hfill
    \begin{minipage}{0.48\linewidth}
    \includegraphics[width=\linewidth]{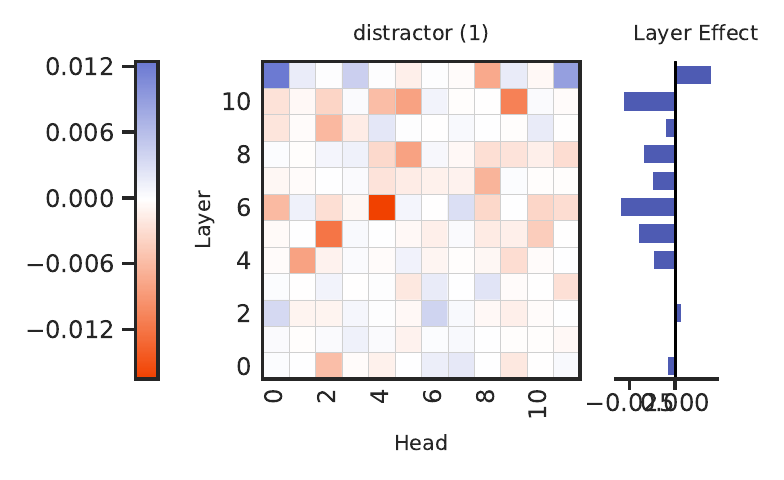}
    \end{minipage}
    \begin{minipage}{0.48\linewidth}
    \includegraphics[width=\linewidth]{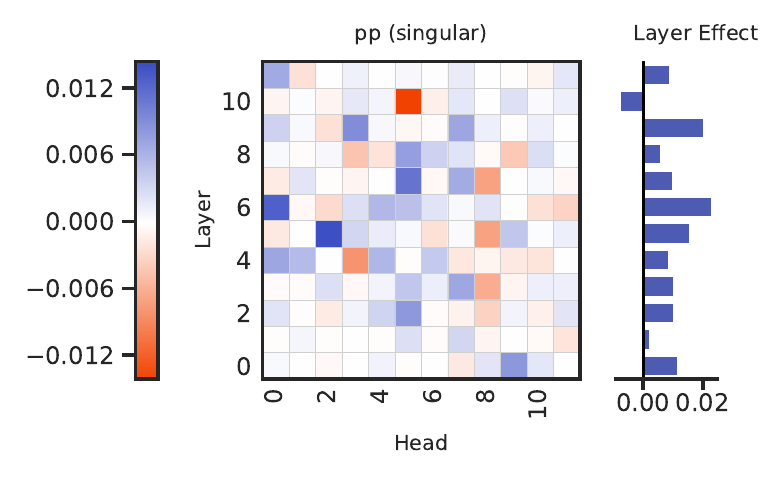}
    \end{minipage}
    \hfill
    \begin{minipage}{0.48\linewidth}
    \includegraphics[width=\linewidth]{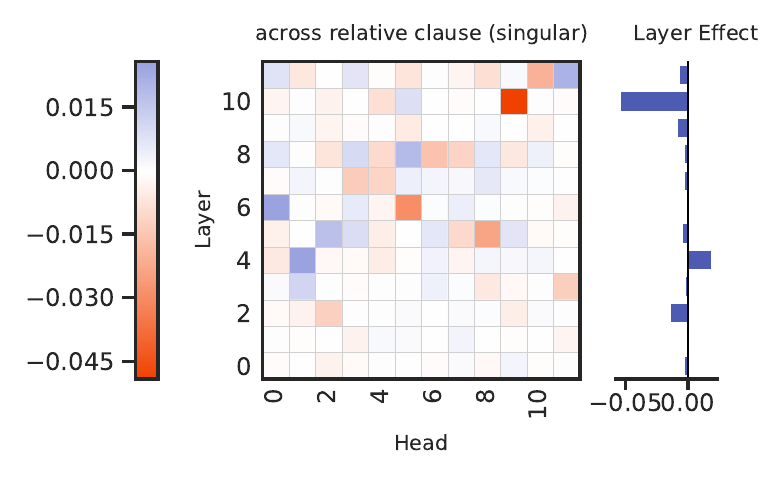}
    \end{minipage}
    \caption{Attention indirect effects for GPT-2 Small under the \texttt{swap-number} intervention.}
    \label{fig:attention_heatmaps_swap}
\end{figure*}

\begin{figure*}
    \centering
    \begin{minipage}{0.48\linewidth}
    \includegraphics[width=\linewidth]{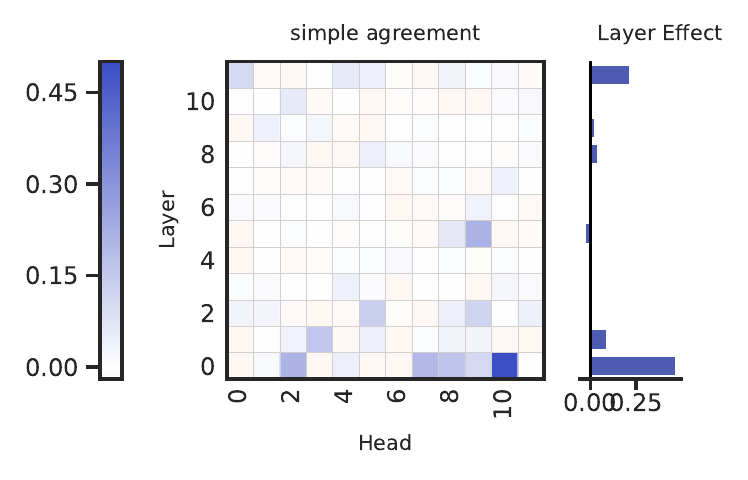}
    \end{minipage}
    \hfill
    \begin{minipage}{0.48\linewidth}
    \includegraphics[width=\linewidth]{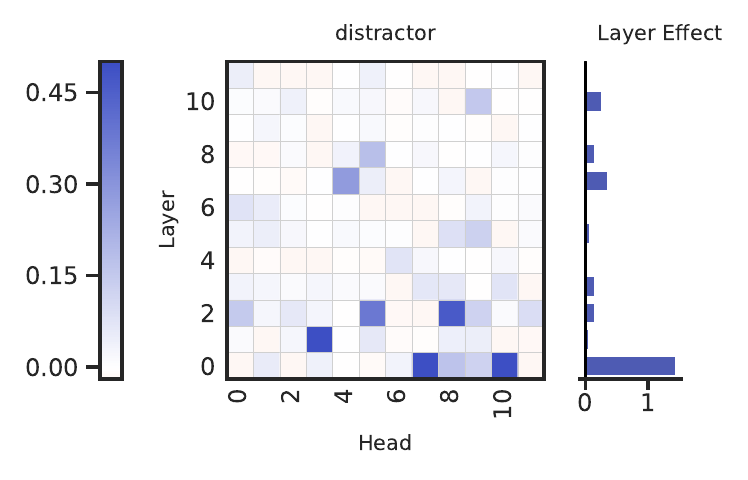}
    \end{minipage}
    \hfill
    \begin{minipage}{0.48\linewidth}
    \includegraphics[width=\linewidth]{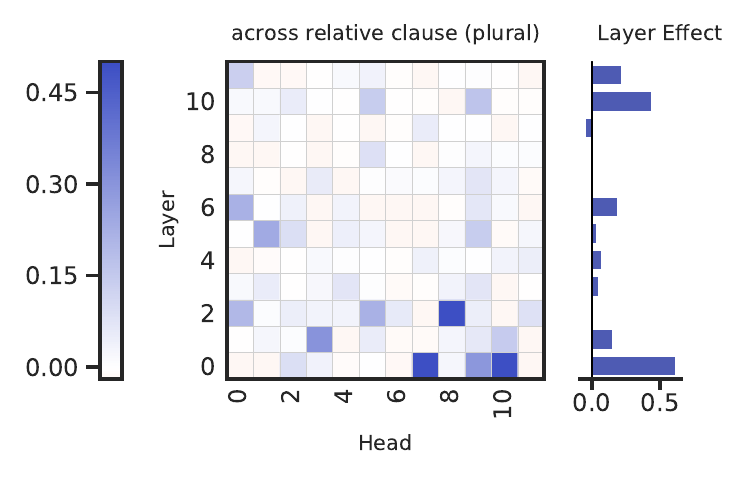}
    \end{minipage}
    \hfill
    \begin{minipage}{0.48\linewidth}
    \includegraphics[width=\linewidth]{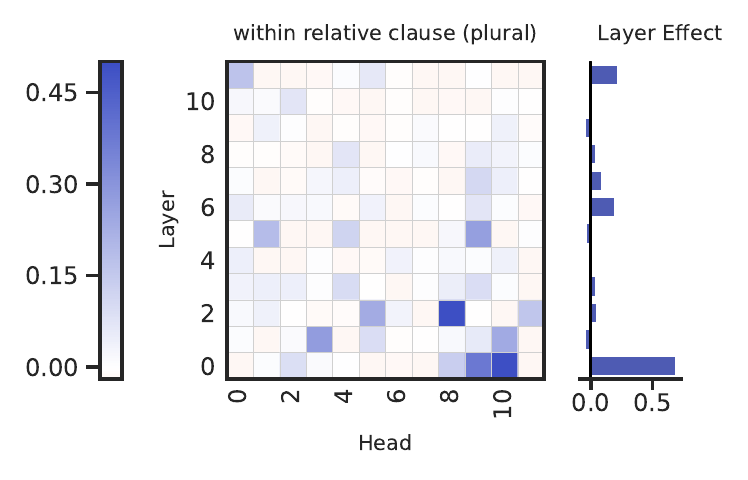}
    \end{minipage}
    \caption{Attention indirect effects for GPT-2 Small under the \texttt{zero} intervention.}
    \label{fig:attention_heatmaps_zero}
\end{figure*}

Here, we present mean indirect effects across prompts for a sample of structures of each attention head in GPT-2 Small (which has 12 layers) under both the \texttt{swap-number} intervention (Figure~\ref{fig:attention_heatmaps_swap}) and \texttt{zero} intervention (Figure~\ref{fig:attention_heatmaps_zero}; defined below).

For the \texttt{swap-number} intervention, we do not observe any consistent trends across structures, except that attention heads in upper-middle layers seem to account for most of the positive \emph{and} negative NIEs. Head 10-9 (layer 10, head 9) has negative indirect effects for most structures where there is a separation between subject and verb, except `across plural RC'. However, we do not observe strong indirect effects for head 10-9 when subject and verb are adjacent. Head 11-11 has the most consistently positive indirect effects across structures, though its magnitude is typically low.

Indirect effects are largely positive for `within plural RC', but otherwise, indirect effects are fairly evenly split between positive and negative. The sum of indirect effects across heads for most structures is close to 0, with many sums being a low-magnitude negative number. This indicates that these attention indirect effects may simply be noise.

Because attention heads seem robust to swapping the number of the subject, we also define the \texttt{zero} intervention. Here, we do not change $u$, but set the attention head's value equal to 0 and observe how this changes the effect; this has an interpretation as the \emph{controlled} indirect effect from \citet{pearl2001direct}. Here, trends are more consistent across structures, attractor numbers, and types of distractors. Head 0-10 is always strongly implicated; since this is in the bottom layer, this suggests that attention's contribution to syntax is based on lexical (perhaps collocational) information and not structural information. This would align with \citet{htut2019attn}, who found that attention tends to capture lexical grammatical features but not inter-word structural information. Qualitative analysis reveals that head 0-10 and head 2-8 always focus on the 2nd and 5th words in the prompt, respectively. Thus, attention's contribution to subject-verb agreement in lower layers may largely be based on where important tokens appear in the input, rather than any abstract structural information that would be composed in the upper layers.

However, for all structures except where we have adverbial distractors, we see consistent positive indirect effects in the uppermost layers as well. No single attention head is strongly implicated, but the layer effect is consistently positive and sometimes nears the magnitude of that in the lower layers. This indicates that more abstract structural information may be present, but that this information is also quite distributed across attention heads in the uppermost layers. Future work should investigate other interventions to better understand attention's role in syntactic agreement.

\section{Adverbs Increase the Probability of Correct Verbs More Than Incorrect Verbs}\label{app:adverbs}
\begin{figure}
    \centering
    \includegraphics[width=\columnwidth]{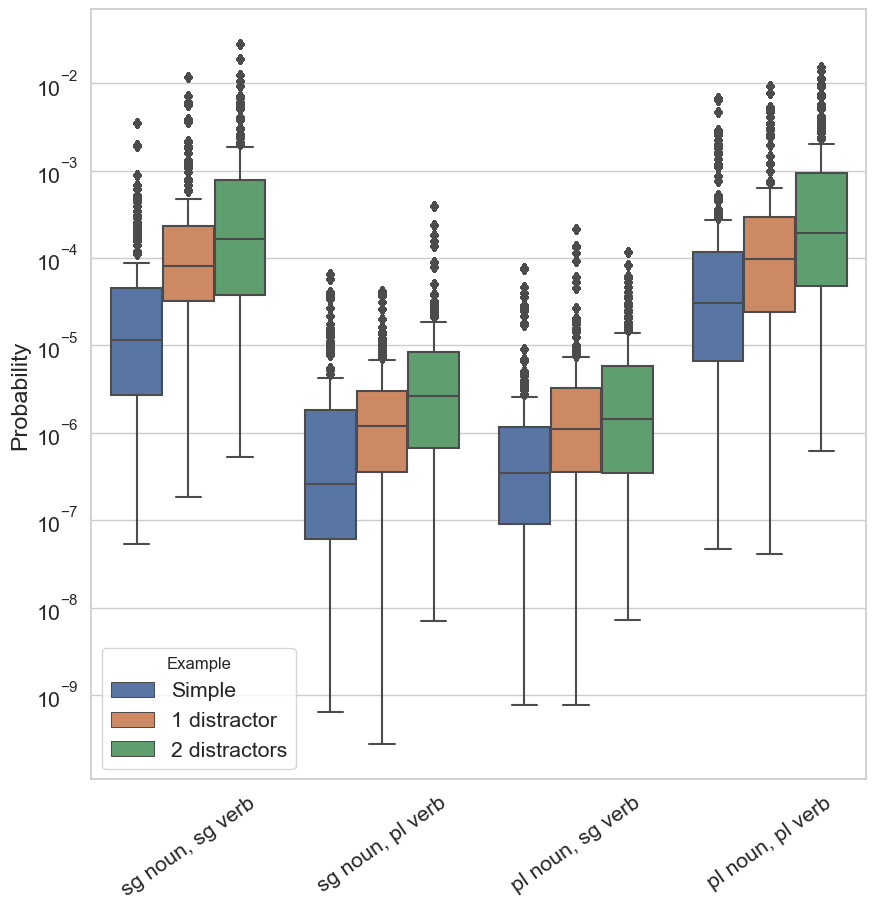}
    \caption{The distribution of target verb probabilities for `simple agreement', `across one distractor', and `across two distractors'. Note that the y-axis uses a log scale.}
    \label{fig:adverbs_increase_prob}
\end{figure}

Here, we show that separating the subject and verb with adverbs tends to increase the probability of the verb (Figure~\ref{fig:adverbs_increase_prob}). Regardless of whether the subject and verb agree, adding adverbs does always increase the probability of verbs. Note the log scale: we observe visually similar increases in log probabilities after adding 1 or 2 adverb distractors, but visually similar differences at higher points in the graph are actually much larger increases. This supports our hypothesis that adverbs increase the probablity of all verbs, but increase the probability of the correct inflection probabilities more than that of the incorrect one.

\section{The (Non-)Impact of Complementizers}\label{app:complementizer}

\begin{figure}[ht]
    \centering
    \includegraphics[width=\columnwidth]{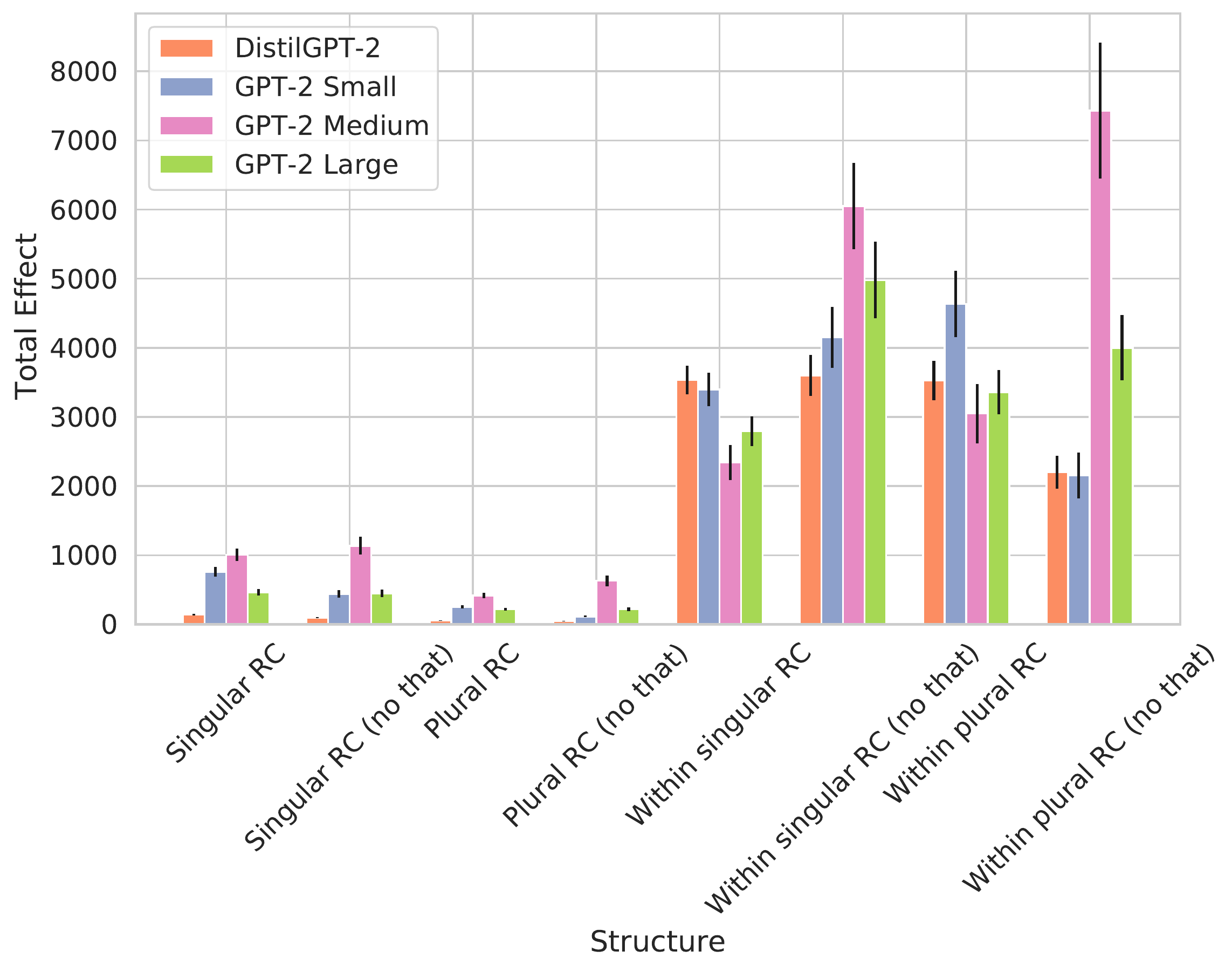}
    \caption{Total effects for `across relative clause' and `within relative clause' structures, with and without the complementizer \emph{that}.}
    \label{fig:total_nocomp}
\end{figure}

\begin{figure}[ht]
    \centering
    \includegraphics[width=\columnwidth]{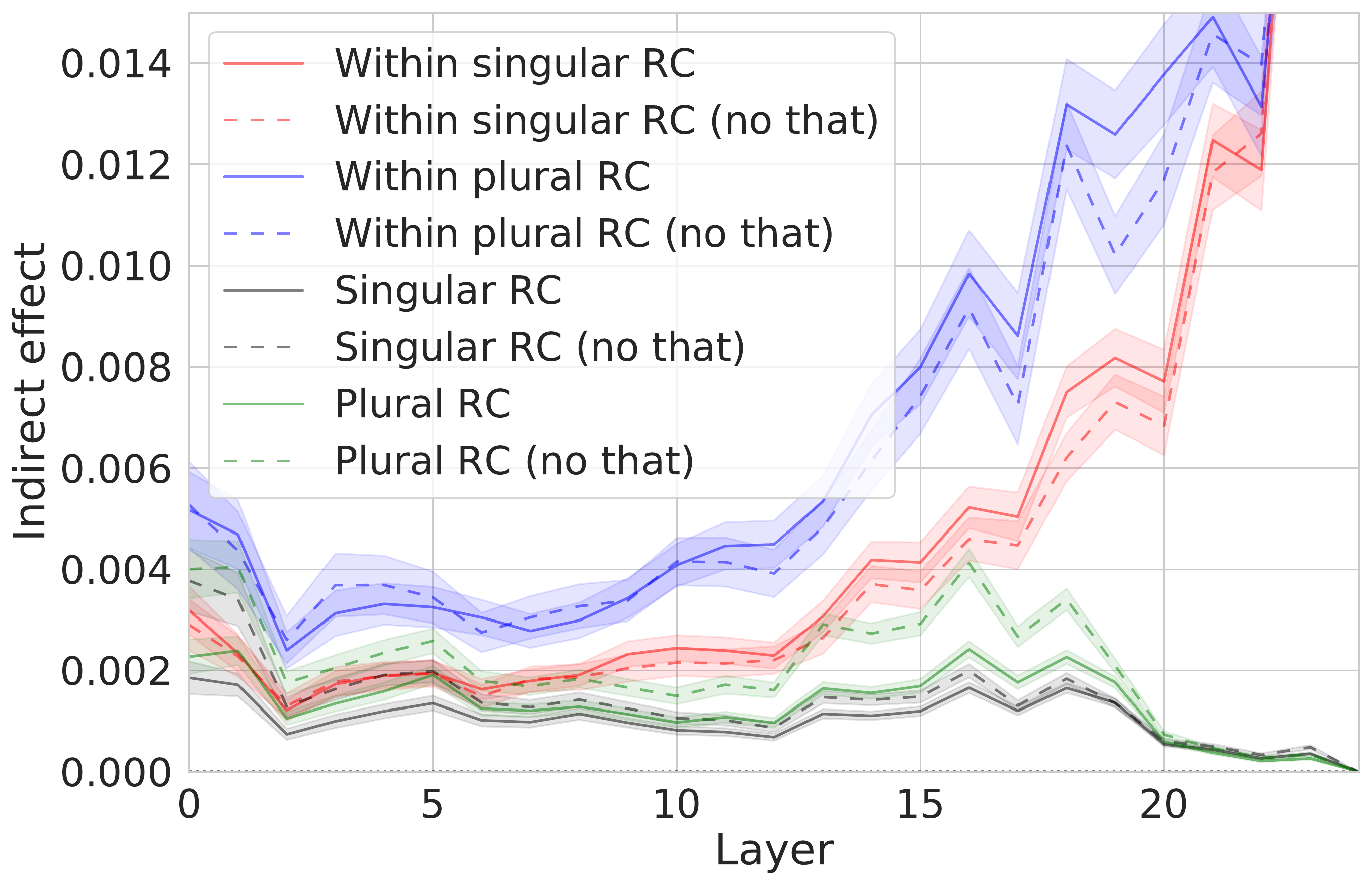}
    \caption{Indirect effects for `across relative clause' and `within relative clause' structures, with (solid lines) and without (dashed lines) the complementizer \emph{that}.}
    \label{fig:neuron_nocomp}
\end{figure}

Here, we investigate the effect of including or excluding the complementizer \emph{that} for the `across RC' and `within RC' structures, observing both TEs (Figure~\ref{fig:total_nocomp}) and neuron indirect effects (Figure~\ref{fig:neuron_nocomp}). While we expect lower TEs when the complementizer is absent, we observe only minor reductions in total effects in `across RC'; this holds across model sizes. For `within RC', however, trends are size-dependent. DistilGPT-2, GPT-2 Small, and GPT-2 Large appear mostly robust to the presence or absence of the complementizer, though GPT-2 Small does have lower total effects in the `across plural RC' structure when \emph{that} is absent. Meanwhile, GPT-2 Medium more strongly prefers correct inflections when \emph{that} is absent. It is not immediately clear why this is the case, because deleting the complementizer introduces more ambiguity.

There does not appear to be any significant difference in magnitude or contour of the indirect effects across layers when including or excluding the complementizer. Thus, while excluding the complementizer can make subject-verb agreement slightly more difficult for LMs~\citep{marvinlinzen18}, it does not appear to change the mechanisms through which subject-verb agreement happens in the model.

\section{Additional Neuron Overlap Details}
\subsection{Hypothesizing Syntactic Similarity}\label{app:hypothesis_matrix}
To generate the hypothesis similarity matrix between structures, we choose a set of features given in Table \ref{tab:sim_features} that capture important syntactic information. Most of the features are binary; however, we also include a ternary feature and a numerical feature. The ternary feature, ``attractor number", can take on values \textsc{sg}, \textsc{pl}, and 0 when there is no attractor. 

To compute the similarity of two structures, we first sum the differences for each feature. For the binary and ternary features, the difference is 0 if the features have the same value, otherwise 1. For the numerical feature, we take the absolute value of the difference between distances, scaled to a value between 0 and 2. We scale the similarity to reduce the impact of the numerical feature on the total similarity.\footnote{We initially scaled this to be within the range [0, 1] like the other features, but this caused ``within relative clause'' to have high similarity to ``across PP'' and ``across a relative clause''. Thus, we increase its impact for more human-like hypotheses.} Finally, we take the maximum possible difference across all pairs of structures, and subtract each pairwise distance from the maximum to obtain similarity scores. We normalize the similarities to the range [0, 100] by dividing similarities by the maximum possible similarity score; this is to make them more comparable to the neuron overlap matrices.

\begin{table}
    \centering
    \resizebox{0.8\columnwidth}{!}{
    \begin{tabular}{ll}
        \toprule
        Feature & Type \\
        \midrule
        Subject and verb separated & binary \\
        Tokens between subject, verb & numerical \\
        Has adverbial distractor(s) & binary \\
        Has noun attractor & binary \\
        Attractor number & ternary \\
        Has relative clause & binary \\
        Has prepositional phrase & binary \\
        \bottomrule
    \end{tabular}}
    \caption{Features (and their types) used in calculating hypothesized syntactic similarity.}
    \label{tab:sim_features}
\end{table}

\begin{figure*}[ht]
    \centering
    \begin{minipage}{0.325\linewidth}
    \includegraphics[width=\linewidth]{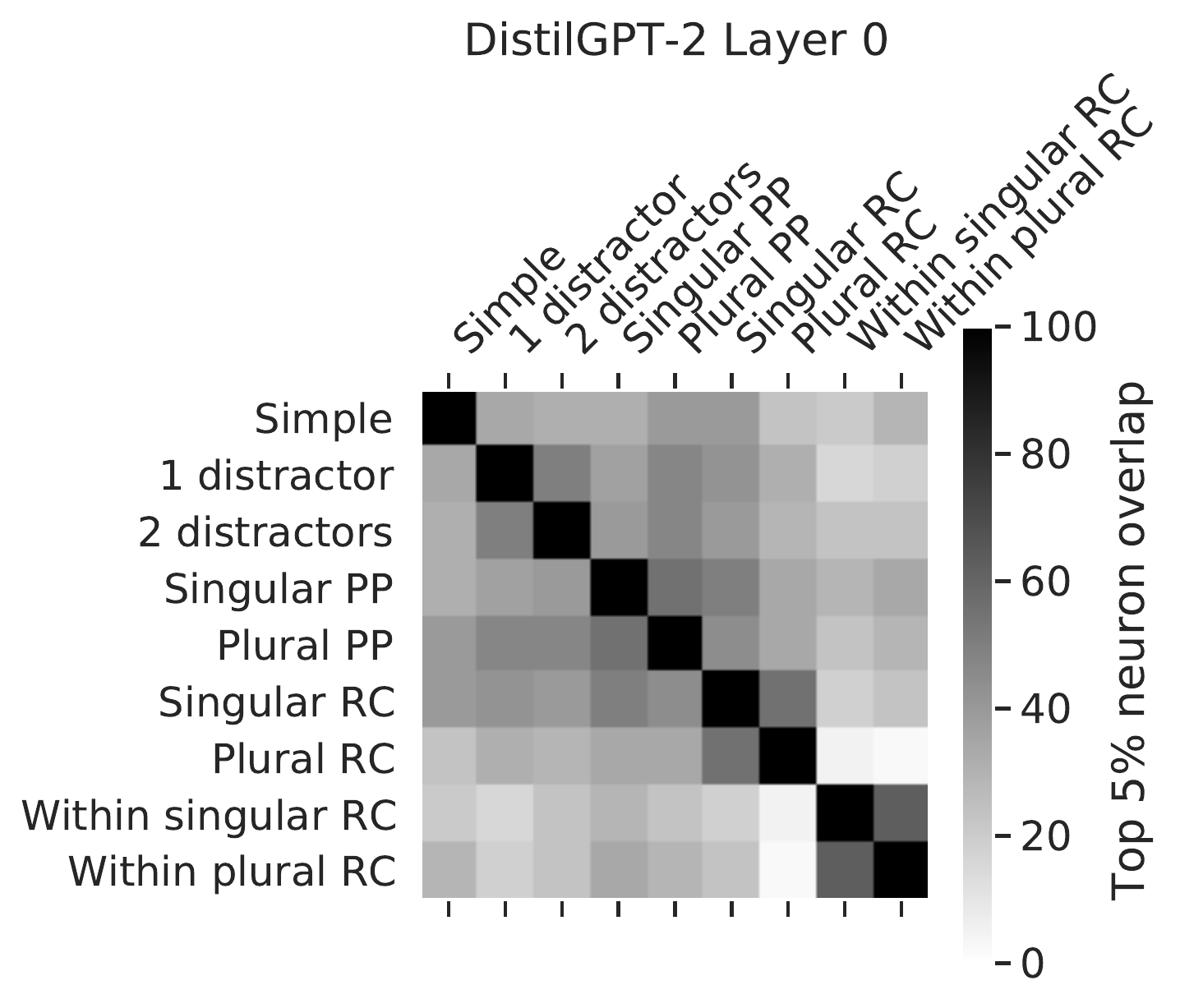}
    \end{minipage}
    \begin{minipage}{0.215\linewidth}
    \includegraphics[trim={4.87cm 0 0 0},clip,width=\linewidth]{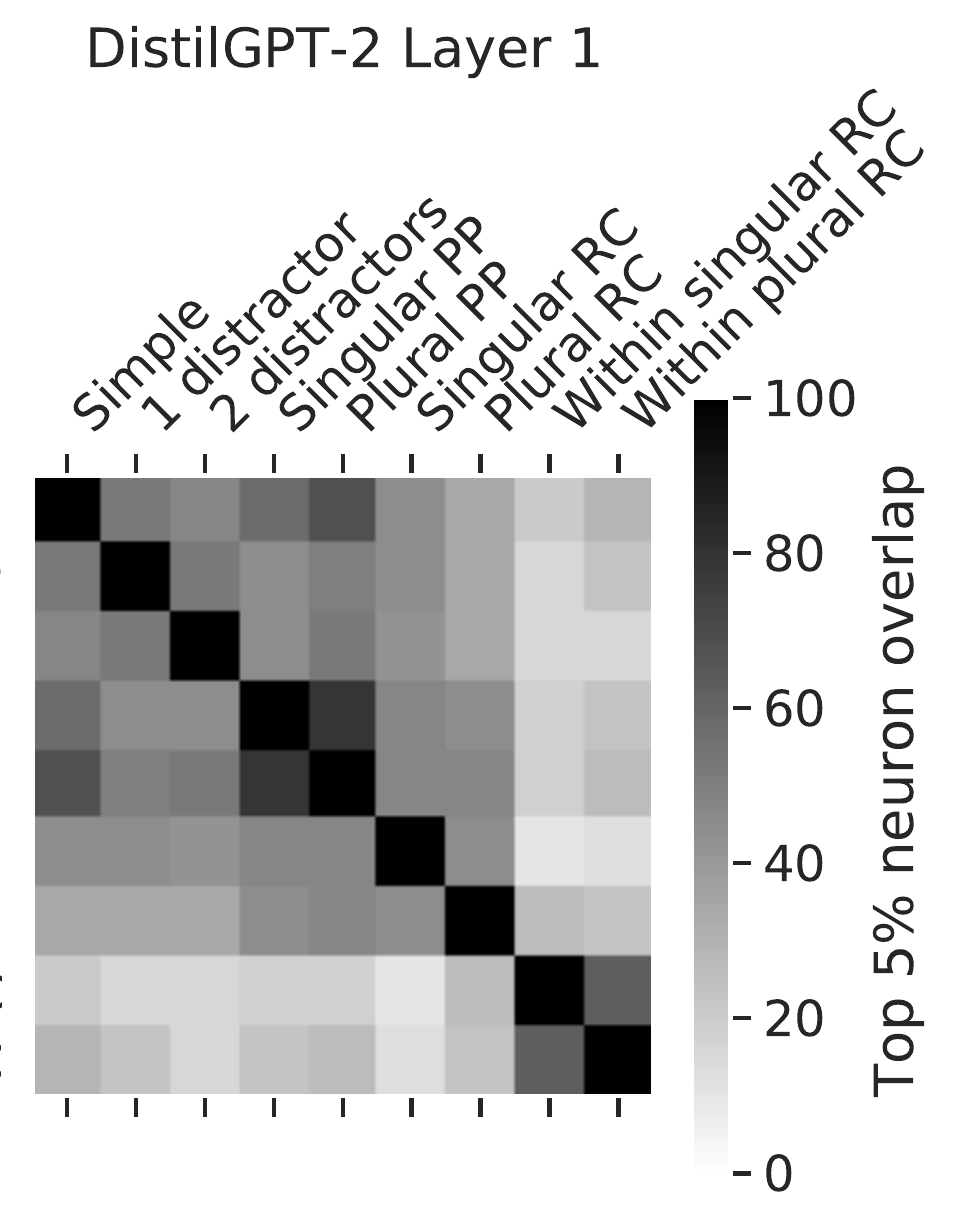}
    \end{minipage}
    \begin{minipage}{0.215\linewidth}
    \includegraphics[trim={4.87cm 0 0 0},clip,width=\linewidth]{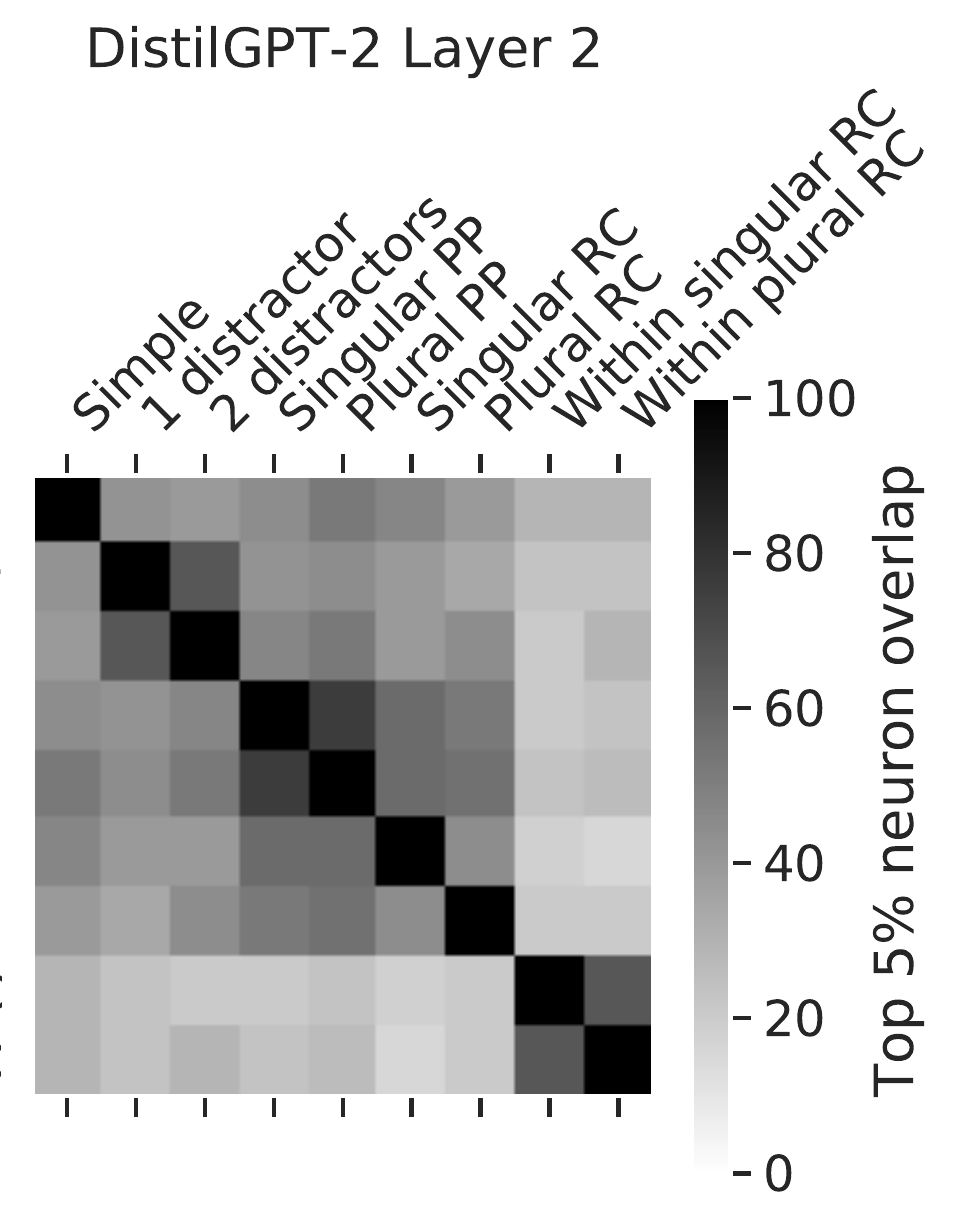}
    \end{minipage}
    \begin{minipage}{0.215\linewidth}
    \includegraphics[trim={4.87cm 0 0 0},clip,width=\linewidth]{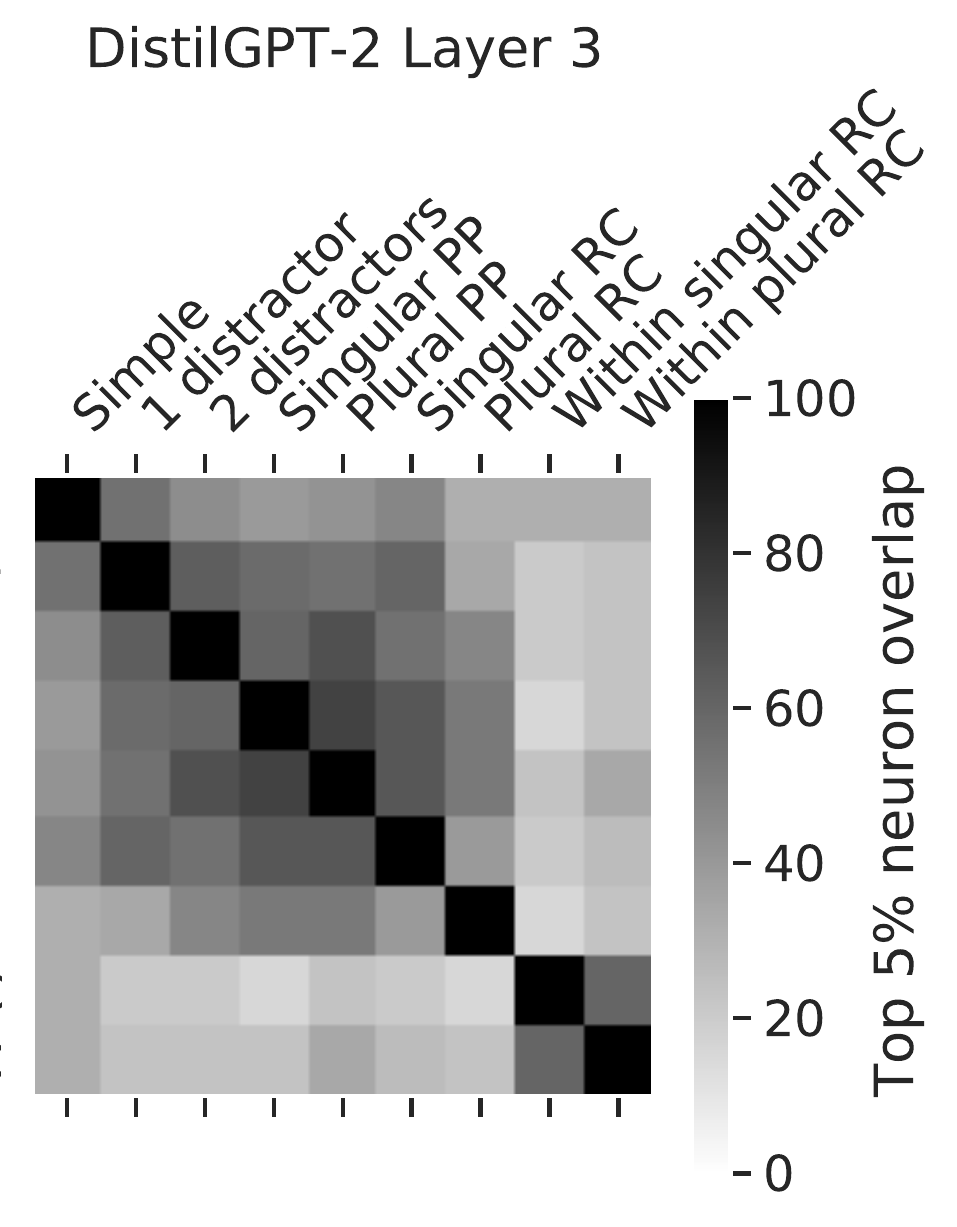}
    \end{minipage}
    \flushleft
    \begin{minipage}{0.325\linewidth}
    \includegraphics[width=\linewidth]{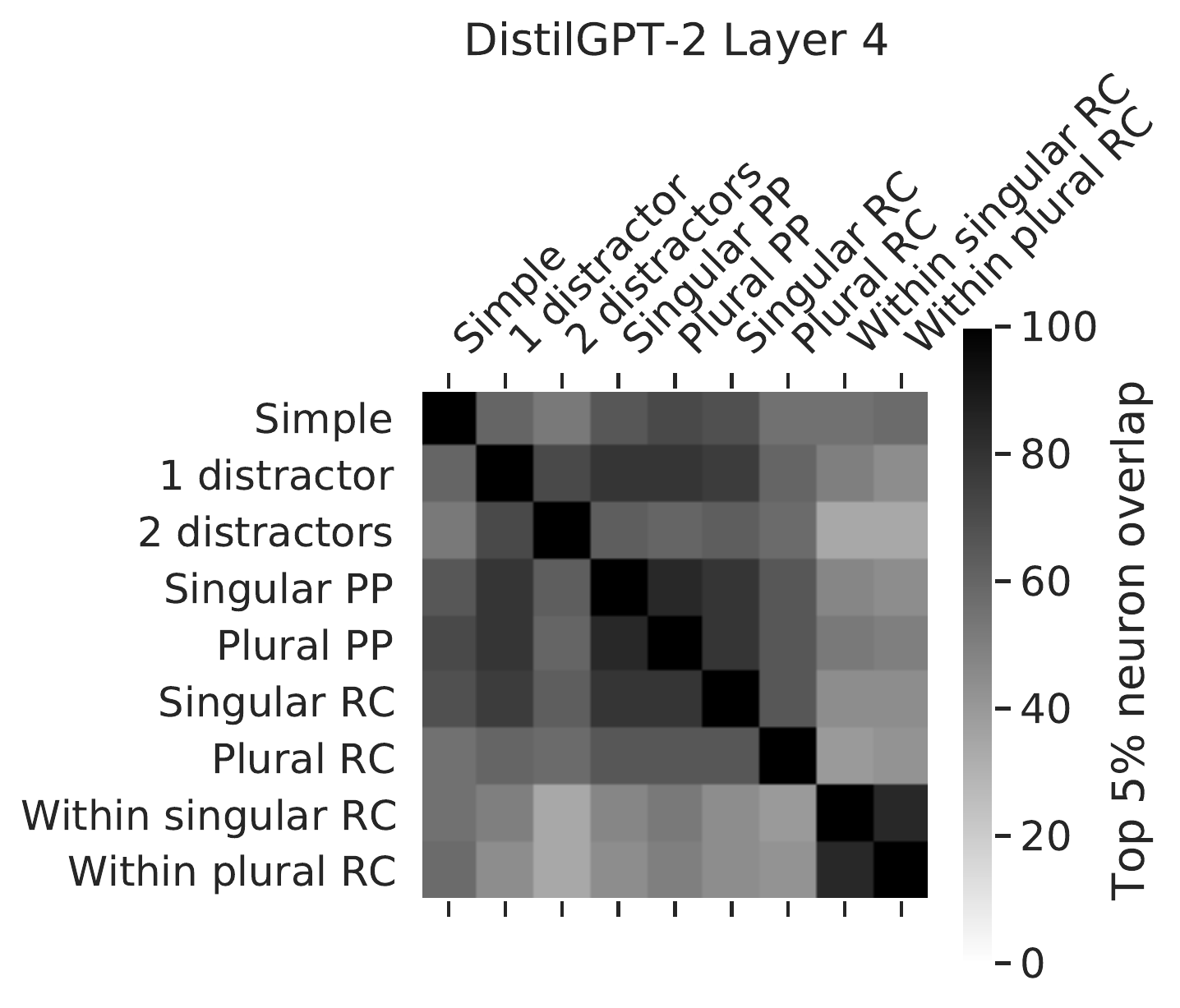}
    \end{minipage}
    \begin{minipage}{0.215\linewidth}
    \includegraphics[trim={4.87cm 0 0 0},clip,width=\linewidth]{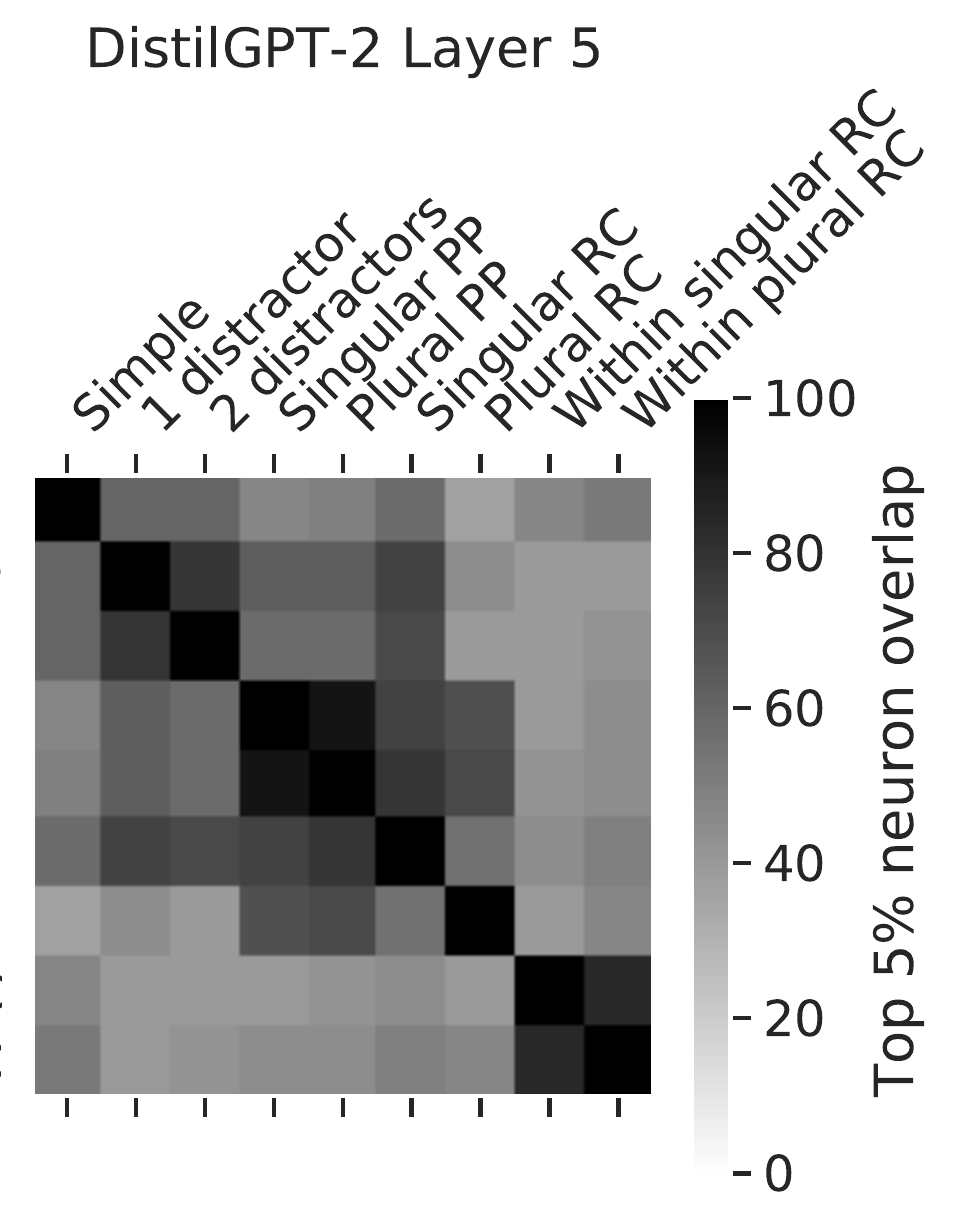}
    \end{minipage}
    \begin{minipage}{0.215\linewidth}
    \includegraphics[trim={4.87cm 0 0 0},clip,width=\linewidth]{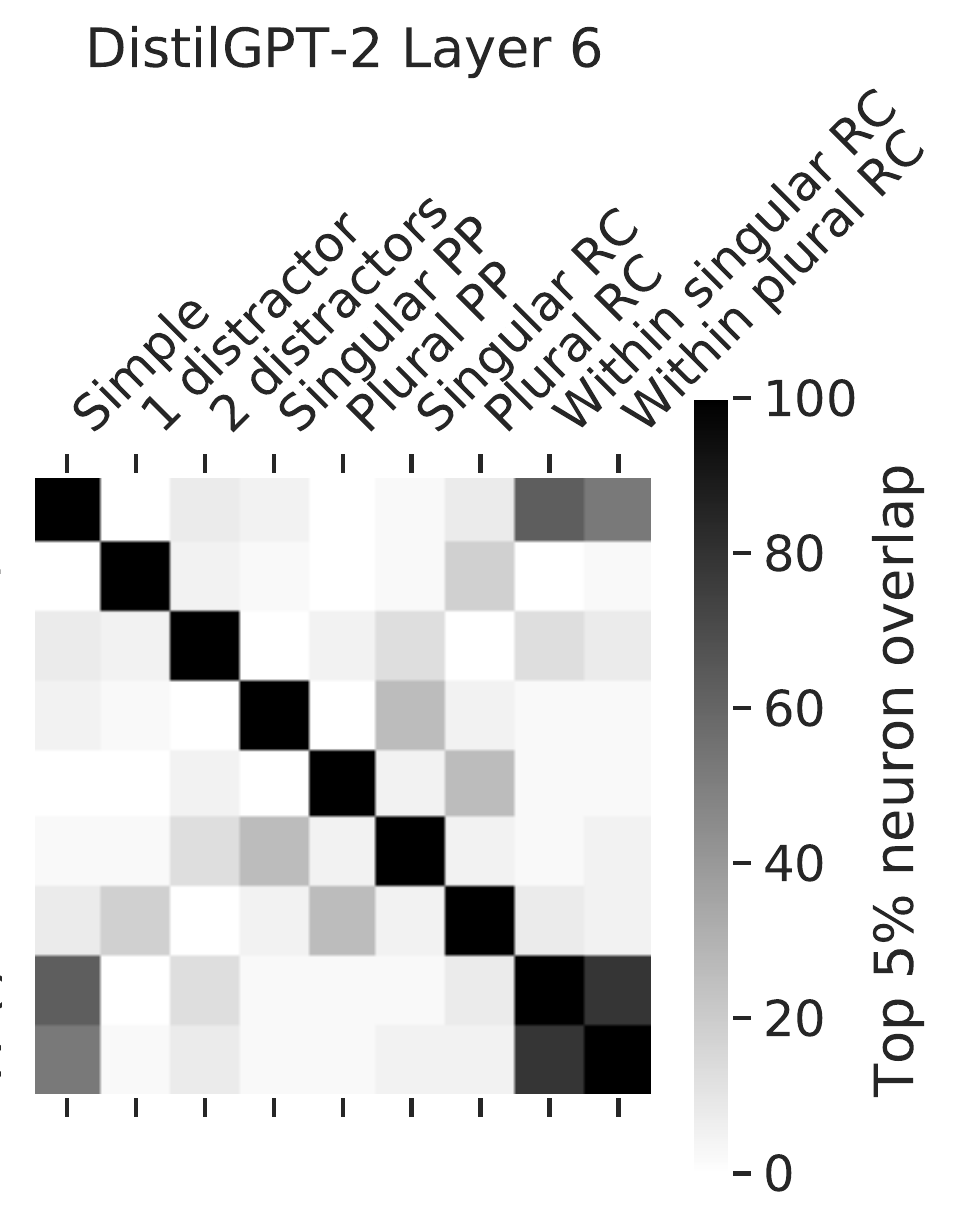}
    \end{minipage}
    \caption{Overlap between structures of the top 5\% of neurons in each of layer of DistilGPT-2 by indirect effect.}
    \label{fig:overlap-by-layer}
\end{figure*}

\subsection{Neuron Overlap Across Layers}\label{app:overlap_by_layer}
Here, we present neuron overlaps across all layers of DistilGPT-2, the smallest model we analyze (Figure~\ref{fig:overlap-by-layer}). We first note that the overall extent of neuron overlap across structures tends to increase up to the upper-middle layers, before sharply decreasing in the highest layer to near-zero values. We find that this trend holds for all other sizes of GPT-2, as well as Transformer-XL; generally, overlaps continue to increase until the upper-middle layers, decreases slightly in the second-highest layer, and decreases sharply to zero in the highest layer. Layer-by-layer difference ($\ell_1$) norms are presented in Table~\ref{tab:diff_norms}.

\begin{table}[h]
    \centering
    \resizebox{0.45\columnwidth}{!}{
    \begin{tabular}[t]{rr}
    \toprule
    Layer No. & Diff.\ Norm \\
    \midrule
    0 & 677 \\
    1 & 652 \\
    2 & 565 \\
    3 & 583 \\
    \bottomrule
    \end{tabular}}
    \resizebox{0.45\columnwidth}{!}{
    \begin{tabular}[t]{rr}
    \toprule
    Layer No. & Diff.\ Norm \\
    \midrule
    4 & 627 \\
    5 & 510 \\
    6 & 1301 \\
    \bottomrule
    \end{tabular}}
    \caption{Difference $\ell_1$ norms between the hypothesis matrix and each layer of DistilGPT-2.}
    \label{tab:diff_norms}
\end{table}

\section{Total Effects Across Architectures}\label{app:te_architectures}

\begin{figure}[h]
    \centering
    \includegraphics[width=\columnwidth]{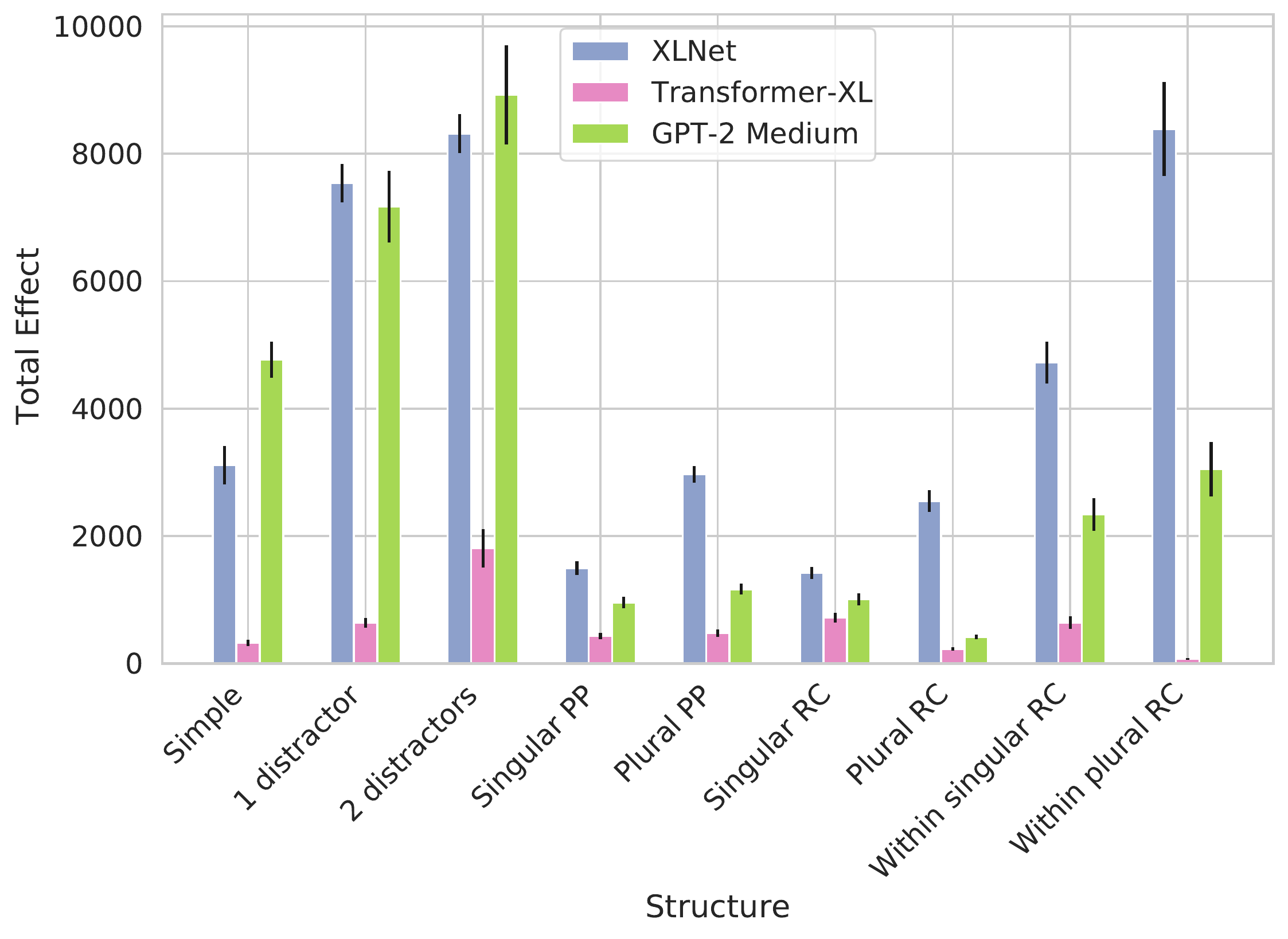}
    \caption{Total effects across structures by architecture (ordered by increasing number of layers/increasing parameterization).}
    \label{fig:te_architectures}
\end{figure}
Figure~\ref{fig:te_architectures} presents total effects for all structures across architectures. The magnitude of total effect is generally similar for XLNet and GPT-2 (except when dealing with relative clauses), whereas total effects are much smaller for Transformer-XL. It seems that parametrization and model depth do not correlate well with total effects.

Perhaps the effects for Transformer-XL are smaller due to the longer effective contexts it has, which could make it prone to assigning smaller probabilities to a larger set of tokens than \mbox{GPT-2} while still behaviorally performing well. It is harder to explain the similarity between GPT-2 and \mbox{XLNet}, given the great differences between them in training and the divergence in their behavior as revealed by the indirect effect results in \S\ref{sec:ie_architectures}.

\end{document}